\documentclass[10pt,journal,compsoc]{IEEEtran}

\usepackage{cite}
\ifCLASSINFOpdf
 
\else
  
\fi

\usepackage[table]{xcolor}
\usepackage{dirtree}
\usepackage{color}
\usepackage{amsmath}
\usepackage{array,graphicx,subfigure}
\usepackage{booktabs}
\usepackage{multirow}
\usepackage{amssymb}
\usepackage{bbding}
\usepackage{tikz}
\definecolor{grn}{rgb}{0,0.7,0}
\usepackage[colorlinks,linkcolor=grn,anchorcolor=grn,citecolor=grn,urlcolor=black,CJKbookmarks=True]{hyperref}
\usepackage{algorithm}
\usepackage{algpseudocode}

\newcommand{\keypoint}[1]{\vspace{0.1cm}\noindent\textbf{#1}}
\newcommand{\tabincell}[2]{\begin{tabular}{@{}#1@{}}#2\end{tabular}}

\newcommand{\cut}[1]{}

\newcommand{\etal}{\textit{et al}.~}
\newcommand{\ie}{\textit{i}.\textit{e}.}
\newcommand{\eg}{\textit{e}.\textit{g}.}
\newcommand{\etc}{\textit{etc}}

\newcommand{\sota}{state-of-the-art~}

\newcommand{\torchsketch}{ \texttt{TorchSketch}}

\begin{document}

\title{Deep Learning for Free-Hand Sketch: A Survey}

\author{Peng~Xu, 
Timothy M. Hospedales,
Qiyue Yin,
Yi-Zhe Song,
Tao Xiang,
and Liang Wang
\IEEEcompsocitemizethanks{\IEEEcompsocthanksitem
This paper is accepted by IEEE TPAMI.\\
Corresponding author: Peng Xu who is with Department of Engineering Science, University of Oxford.\\
E-mail: peng.xu@eng.ox.ac.uk,\\ HomePage: \url{http://www.pengxu.net/}
}
}

\IEEEtitleabstractindextext{
\begin{abstract}
Free-hand sketches are highly illustrative, and have been widely used by humans to depict objects or stories from ancient times to the present.
The recent prevalence of touchscreen devices has made sketch creation a much easier task than ever and consequently made sketch-oriented applications increasingly popular.
The progress of deep learning has immensely benefited  free-hand sketch research and applications.
This paper presents a comprehensive survey of the deep learning techniques oriented at free-hand sketch data, and the applications that they enable. 
The main contents of this survey include:
(i) A discussion of the intrinsic traits and unique challenges of free-hand sketch, to highlight the essential differences between sketch data and other data modalities, \eg, natural photos.
(ii) A review of the developments of free-hand sketch research in the deep learning era, by surveying existing datasets, research topics, and the state-of-the-art methods through a detailed taxonomy and experimental evaluation.
{(iii) Promotion of future work via a discussion of bottlenecks, open problems, and potential research directions for the community.
}
\end{abstract}

}
\maketitle
\IEEEdisplaynontitleabstractindextext
\IEEEpeerreviewmaketitle

\IEEEraisesectionheading{\section{Introduction}\label{sec:introduction}}

\begin{figure*}[!t]
\begin{center}
\includegraphics[width=0.95\textwidth]{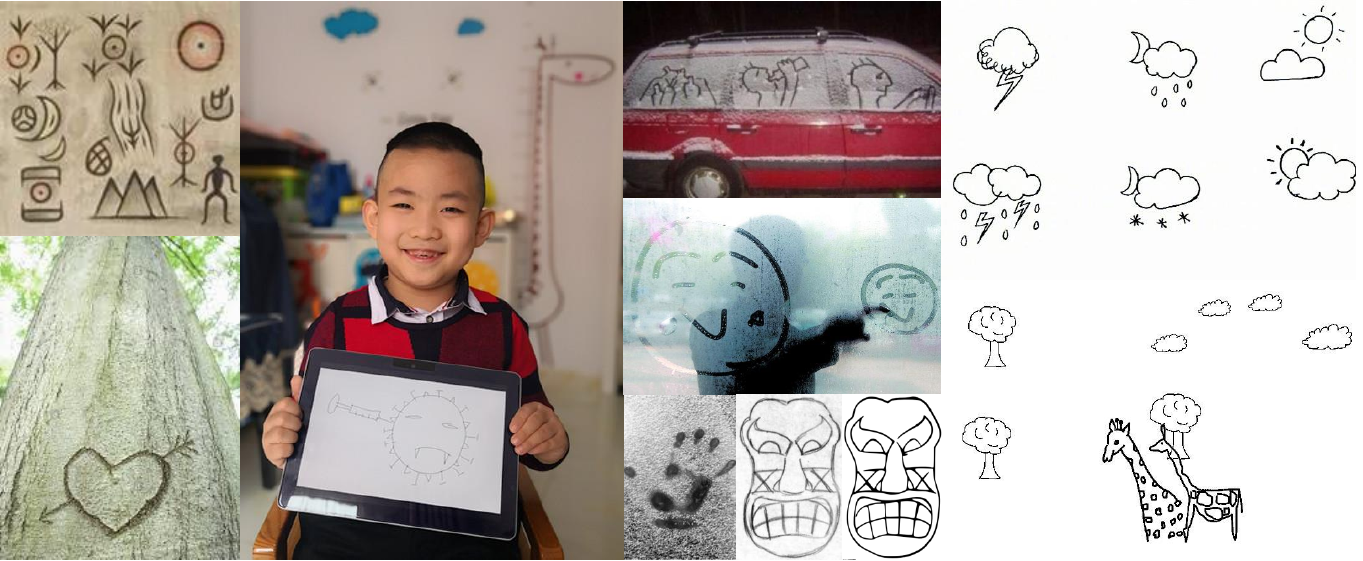}
\end{center}
   \caption{Diverse free-hand sketches in human daily life.  {The masks (rough and simplified) on the bottom are from \cite{simo2016learning}.} {The scene-level sketch (cloud, trees, and giraffes) on the bottom right corner is from SketchyCOCO dataset~\cite{Gao2020SketchyCOCO}.}
   }
\label{fig:sketches}
\end{figure*}

\IEEEPARstart{F}{ree-hand} sketch is a universal communication and art modality that transcends barriers to link human societies. It has been used from ancient times to today, comes naturally to children before writing, and transcends language barriers. Different from other related forms of expression such as professional sketch, forensic sketch, cartoons, technical drawing, and oil paintings, it requires no training and no special equipment. As such free-hand sketch is not bound by age, race, language, geography, or national boundaries. It can be regarded as an expression of the brain's internal representation of the world, whether perceived or imagined.  Smiling faces, for example, are always recognized by humans (Figure~\ref{fig:sketches}). 

Sketches can convey many words, or even concepts that are hard to convey at all in words. Figure~\ref{fig:sketches} shows several examples covering ancient and contemporary; literal and emotional; iconic and descriptive; abstract and concrete; and different media of drawing. 

Free-hand sketch can be illustrative, despite its highly concise and abstract nature, making it useful in various scenarios such as communication and design. Therefore, free-hand sketch has been widely studied in computer vision and pattern recognition ~\cite{hu2013performance,Wang_2015_CVPR,yu2017sketch,Yu_2016_CVPR,ha2018sketchrnn}, computer graphics \cite{eitz2012humans,sangkloy2016sketchy}, human computer interaction~\cite{huang2019swire,suleri2019eve,kwan2019mobi3dsketch,gasques2019you}, robotics \cite{kotani2019robotDraw}, and cognitive science \cite{fan2019collabdraw} communities.  In particular, early research can be traced back to the 1960s and 1970s~\cite{sutherland1964sketchpad,herot1976graphical}.  

However, free-hand sketch is fundamentally different from natural photos\footnote{Images can include both free-hand sketch and natural photos, \etc. In this survey, ``photo'' denotes natural photo images obtained by a camera such as in ImageNet unless otherwise specified.}. 
Sketch images provide a special data modality/domain that has both domain-unique challenges (\eg, highly sparse, abstract, artist-dependent) and advantages (\eg, lack of background, use of iconic representation). It is also unique in that free-hand sketch can be stored and processed in multiple representations as its source is a dynamic `pen' movement. These include static pixel space (when rendered as an image), dynamic stroke coordinate space (when considered as a time series), and geometric graph space (when considered topologically) -- as discussed in  Section~\ref{sec:background}. Thus, from a pattern recognition or machine intelligence perspective, these unique traits of free-hand sketch often lead to sketch-specific model designs in order to exploit sketch-specific data properties and overcome sketch-specific challenges when analyzing sketches for recognition, generation, and so on. This survey will review these considerations and designs in detail.

\begin{figure}[!t]
\begin{center}
\includegraphics[width=\columnwidth]{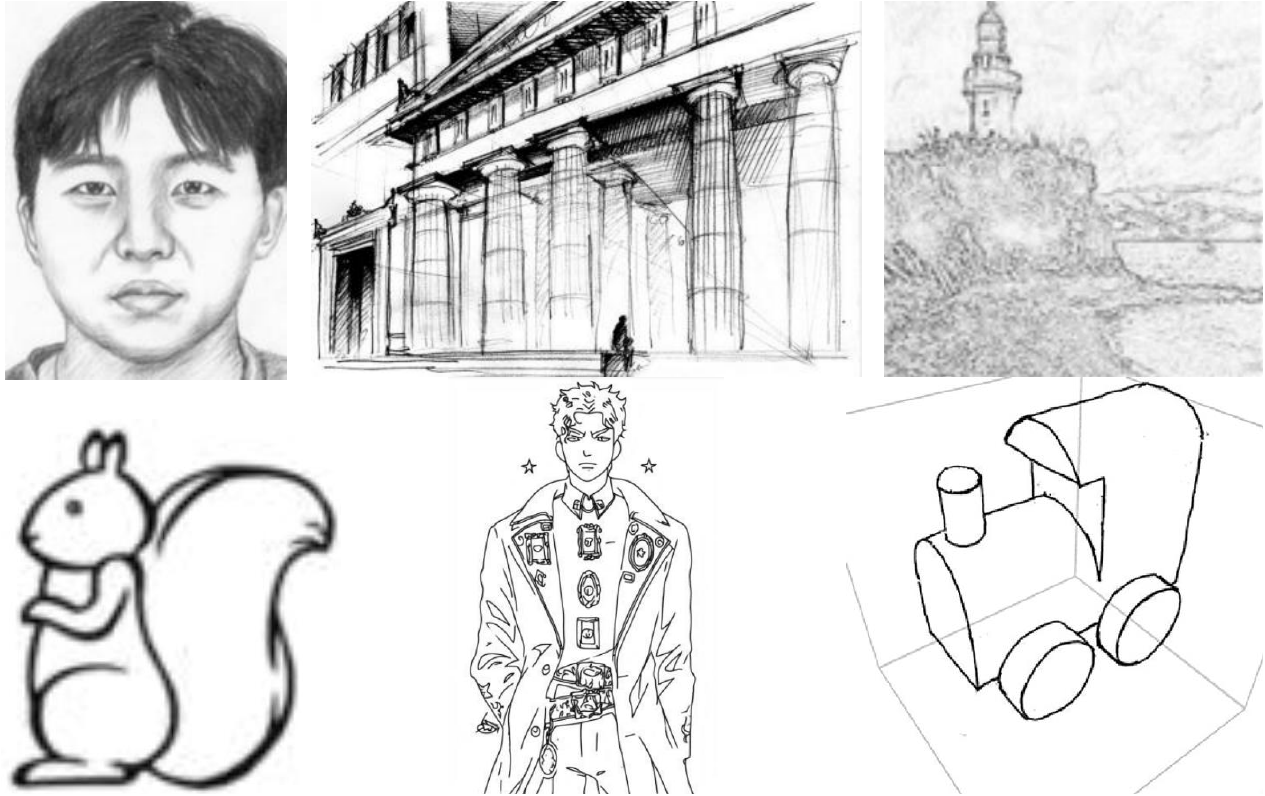}
\end{center}
   \caption{{Drawing samples out-of-scope of our focus on free-hand sketch.}}
 
\label{fig:out-of-scope}
\end{figure}

Sketch research and applications in both industry and academia have boomed in recent years due to the prevalence of touchscreen devices (\eg, smartphone, tablet) that make acquiring sketch data much easier than ever; as well as the rapid development of deep learning techniques that are achieving 
\sota performance in diverse artificial intelligence tasks.
This boom has occurred on several fronts:
(i) Some classic research topics (\eg, sketch recognition, sketch-based image retrieval, sketch-based 3D shape retrieval) have been re-studied in a deep learning context~\cite{yusketch2015BMVC,yu2017sketch,sangkloy2016sketchy,Yu_2016_CVPR,Wang_2015_CVPR,xu2018sketchmate} resulting in significant performance improvements.
(ii) Some brand-new topics have been proposed based on deep learning, \eg, deep learning based sketch generation/synthesis~\cite{ha2018sketchrnn}, sketch-based model generation~\cite{Hu_2018_CVPR},  reinforcement learning based sketch abstraction~\cite{muhammad2018learning},
adversarial sketch based image editing~\cite{portenier2018faceshop},
graph neural network based sketch recognition~\cite{xu2019multi},
graph convolution-based sketch semantic segmentation~\cite{yang2020sketchgcn}, 
and 
sketch based software prototyping~\cite{suleri2019eve}.
(iii) Beyond global representation based tasks (\eg, sketch recognition), more instance-level and stroke-level tasks have been further studied or proposed, \eg, instance-level sketch-based image retrieval~\cite{Yu_2016_CVPR}, and deep stroke-level sketch segmentation~\cite{qi2019sketchsegnet+}.
(iv) Compared with the conventional approach of representing sketches as static images \cite{hu2013performance}, the trends of touchscreen acquisition and deep learning have underpinned progress on designing deep network architectures to exploit richer representations of sketch. Thanks to works such as SketchRNN~\cite{ha2018sketchrnn}, the sequential nature of free-hand sketches is now widely modeled by recurrent neural network (RNN).
(v) More sketch-based applications have appeared, \eg, the online sketch game QuickDraw\footnote{\url{https://quickdraw.withgoogle.com}}~\cite{ha2018sketchrnn}, and sketch-based commodity search engine\footnote{\url{http://sketchx.eecs.qmul.ac.uk/demos/}}~\cite{Yu_2016_CVPR,Song_2017_ICCV}.
(vi) Some large-scale sketch datasets have been collected, \eg, Sketchy \cite{sangkloy2016sketchy} and Google QuickDraw\footnote{\url{https://github.com/googlecreativelab/quickdraw-dataset}}~\cite{ha2018sketchrnn} -- a million-scale sketch dataset (50M+).

\subsection{Overview}

This survey aims to review the state of the free-hand sketch community in deep learning era, hoping to bring insights to researchers and assist practitioners aiming to build sketch-based applications.  

\keypoint{Previous Surveys and Scope}
This survey focuses on free-hand sketches, not including
professional (forensic) facial sketch~\cite{Ouyang_2016_CVPR,hu2017now,Nagpal_2017_ICCV,Fan_2019_ICCV,pang2018cross}, professional pencil sketch ({professional line drawing/art})~\cite{huang2018reversed,SasakiCGI2018,zheng2020learning,Lee2020Colorization,gui2020learning,yan2020benchmark,ZhangCVPR2021}, 
professional landscape sketch~\cite{chen2019sragan}, photo-like edge-maps (artificially rendered `sketch')~\cite{amer2015monocular,wang2019learning,chen2019human}, cartoon/manga~\cite{han2018caricatureshop,zhao2019cartoonish,YuanCVPRW2020},
well-drawn 3D sketch~\cite{delanoy20183d}. {(See Figure~\ref{fig:out-of-scope}.)}
In this survey, ``sketch'' refers to ``free-hand sketch'', unless otherwise specified.

To our knowledge, only a few survey papers~\cite{li2014comparison,prajapati2015sketch,indu2016survey,li2018survey,zhang2019survey} were published in the free-hand sketch community in the past decade. However, these survey papers~\cite{li2014comparison,prajapati2015sketch,indu2016survey,li2018survey,zhang2019survey}: 
(i) only focus on two research topics, \ie, free-hand sketch based recognition and image/3D retrieval. (ii) mainly review classic non-deep techniques. In contrast, the current boom in advanced deep methodologies, techniques (hashing), representations (sequential, topological), and novel applications (generation, segmentation) makes it timely to provide an up-to-date survey of the big picture of research on free-hand sketch. 



\keypoint{Contributions}
We provide a comprehensive survey reviewing the state of the field with regards to deep learning techniques, as well as applications of free-hand sketch. In particular: (a) We discuss the intrinsic traits, and unique challenges and opportunities posed when working with free-hand sketch data. (b) We provide a detailed taxonomy of both datasets, and applications covering both uni-modal (sketch alone) and multi-modal (relating sketches to photos, text, \etc) cases. For each specific task, the contemporary landscape of deep learning solutions is summarized, and milestone works are described in detail. (c) We discuss current bottlenecks, open problems, and potential research directions for free-hand sketch.

\keypoint{Organization of This Survey} 
The rest of this survey is organized as follows.
Section~\ref{sec:background} provides background  on free-hand sketch, including intrinsic traits, domain-unique challenges, milestone techniques of the existing sketch-oriented deep learning works, \etc. Section~\ref{sec:datasets} summarizes representative free-hand sketch datasets. In Section~\ref{sec:taxonomy}, we provide a comprehensive taxonomy for various sketch-based tasks, and describe  representative deep learning techniques in detail. 
Section~\ref{sec:taxonomy} also presents some experiment comparisons based on~\torchsketch\footnote{An open source sketch-oriented deep learning software library. Please see its GitHub page for details \url{https://github.com/PengBoXiangShang/torchsketch}.}~implementation.
Section~\ref{sec:thinking_and_discussion} discusses open problems, bottlenecks, and potential research directions before the survey concludes in Section~\ref{sec:conclusion}.

Throughout this survey,
bold uppercase and bold lowercase characters denote matrices and vectors, respectively.
Unless specified otherwise, mathematical symbols and abbreviated terms follow the conventions in Table~\ref{table:definitions}.
\begin{table}[!t]

\caption{Notation and abbreviations used in this survey.}
\label{table:definitions}
\begin{center}
\resizebox{\columnwidth}{!}{
\begin{tabular}{ l | l }
\hline
Notations & Descriptions \\
\hline
$\mathcal{X} = \{{\bf X}_{n}\}_{n=1}^{N}$ & sketch sample set \\
${\bf X}_{m}$, ${\bf X}_{n}$ & $m$-th and $n$-th sketch samples in the sketch sample set $\mathcal{X}$ \\
$\mathcal{Y} = \{y_{n}\}_{n=1}^{N}$ & associated label set of $\mathcal{X}$\\
$y_{n}$ & label of ${\bf X}_{n}$ \\
$\mathcal{L}$ & loss function \\
${\bf \Theta}$ & learnable parameters of neural network \\ 
$\mathcal{F}(\cdot)$ & function mapping or feature extraction \\ 
$\mathcal{F}_{\Theta}(\cdot)$ & neural network feature extraction, parameterized by $\Theta$ \\ 
$\mathcal{D}(\cdot, \cdot)$ & distance metric, \eg, $\ell_{2}$ distance \\ 
$\lambda$ & weighting factor \\
$\sum$ &  summation \\
$\alpha$, $\beta$, $\gamma$  & {hyper parameters set manually} \\

\hline
\hline
\tabincell{c}{Abbreviated \\ Terms} & Descriptions \\
\hline
CNN &  convolutional neural network \\
GNN & graph neural network \\
{GCN} & {graph convolutional network} \\
RNN & recurrent neural network \\
{LSTM} & {Long Short Term Memory} \\
{GRU} & {Gated Recurrent Unit} \\
{BERT} & {Bidirectional Encoder Representations from Transformers} \\
TCN & temporal convolutional neural network \\
{GAN} & {generative adversarial network}    \\
{VAE} & {Variational Auto Encoder} \\
{RL} & {Reinforcement Learning} \\
\hline
\end{tabular}
}
\end{center}

\end{table}

\section{Background}
\label{sec:background}

This section presents background knowledge, including: the intrinsic traits, and domain-unique challenges and opportunities of free-hand sketch. In particular, we cover  the essential differences to natural photos; and 
a brief development history of deep learning for free-hand sketch, summarizing the milestone techniques. 

\subsection{{Intrinsic Traits and Domain-Unique Challenges}}
\label{sec:Intrinsic_Traits_and_Domain-Unique_Challenges}

\begin{figure}[!t]
\begin{center}
\includegraphics[width=\columnwidth]{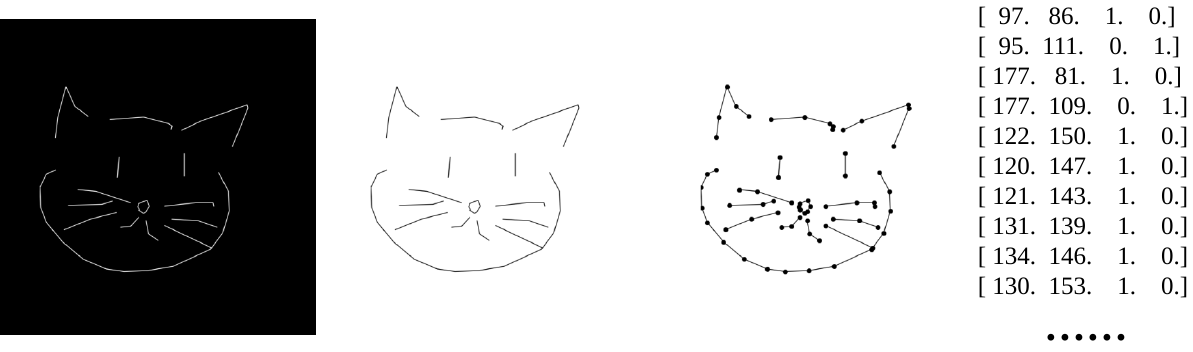}
\end{center}
   \caption{Sketch-specific representations. Representations from left to right: sparse matrix (black background with white lines), dense picture (white background with black lines), graph, stroke sequence. Both graph and stroke sequence representations are based on the key stroke points. In stroke sequence, each key point is denoted as a four-tuple, where the first two entries and the last two entries represent the coordinates and pen state, respectively. See details in text.}
\label{fig:motivations}
\end{figure}

\begin{figure}[!t]
\begin{center}
\includegraphics[width=\columnwidth]{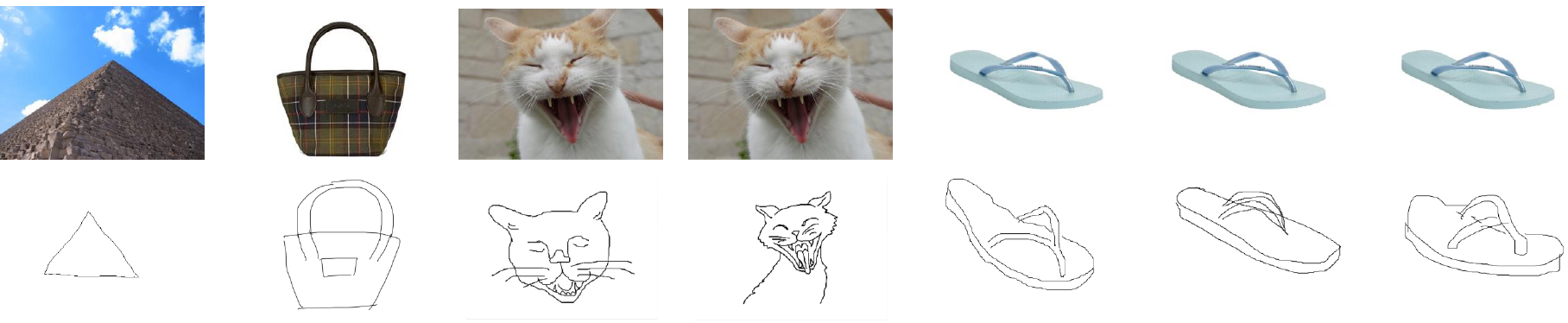}
\end{center}
   \caption{Illustrations of the major domain-unique challenges of free-hand sketch. Each column is a photo-sketch pair. Sketch is highly abstract. A pyramid can be depicted as a triangle in sketch, and a few strokes depict a fancy handbag.  Sketch is highly diverse. Different {people} draw distinctive sketches when given the identical reference, due to subjective salience (head vs. body), and drawing style.}
\label{fig:sketch_unique_challenges}
\end{figure}


\keypoint{Representation}
Free-hand sketch is a special kind of visual data, intrinsically different to natural photos that are the pixel-perfect copies of the real world. For efficient storage and fast calculation, free-hand sketch can be saved as a sparse matrix, or as a black and white image that ignores its sparsity (Figure~\ref{fig:motivations}, left images). Since sketch generation is a dynamic process, suitably captured sketches can also be represented as a sequence of strokes or pen coordinates (Figure~\ref{fig:motivations}, right). In this regard, sketches share similarities with hand-written characters, yet are fundamentally different given their highly abstract and free-style nature (c.f., alphabetic hand-writing is subject to specific rules and a teaching process). 
From another perspective, free-hand sketches can also be modeled as a sparsely connected graph where lines are edges in the graph. Compared with a sequence of Euclidean coordinates, topological representation as a graph can provide a more flexible and abstract representation. As a result of this diversity of possible representations, various deep learning paradigms can be used to process sketches including CNNs, RNNs, GCNs, and TCNs.


\keypoint{Unique Challenges and Opportunities}
The unique challenges of free-hand sketch can be summarized as follows: 
(i) Abstraction: Humans use sketch to depict an object or event in very few strokes, reflecting the high-level semantics of a mental image. As shown in Figure~\ref{fig:sketch_unique_challenges}, a pyramid can be depicted as a triangle in sketch, and few strokes can depict a fancy handbag.
(ii) Diversity: Different {people} have different drawing styles. For example, a near `realistic' (close to photo edge-map) sketch image could be portrayed in different ways as exaggerated (c.f. caricatures), iconic (where details are omitted and the sketch is near symbolic), or artistic. Depending on subjective opinion about salience, different parts may also be included or omitted in a sketch. For instance, given a concept ``cat'', people differ on choice of drawing with/without body (Figure~\ref{fig:sketch_unique_challenges}). Finally, there is the mental viewpoint of different users, \eg, whether they imagine an orthographic or perspective projection image.
In Figure~\ref{fig:sketch_unique_challenges}, we can see that different {people} draw differing perspective views of an identical slipper. 
(iii) Sparsity: No matter the representation, free-hand sketch is a highly sparse signal compared to photographs. {(iv) Invariance: People can still recognize sketches after they are shifted, rescaled, rotated, or flipped. In Section~\ref{sec:robustness-study}, we conduct a robustness study to evaluate whether deep networks are sensitive to spatial transformations in sketch related tasks.}
(v) Finally, there are  some unique challenges when collecting sketch, which will be discussed in detail as follows (see Section~\ref{sec:Challenge_of_Sketch_Collection}).


\begin{figure}[!th]
	\centering
	\subfigure[ A sketch is drawn with the device on the back of the hand~\cite{Schrapel2020WatchMP}.]{
		\label{fig:watch-my-painting}
		\includegraphics[width=0.8\columnwidth]{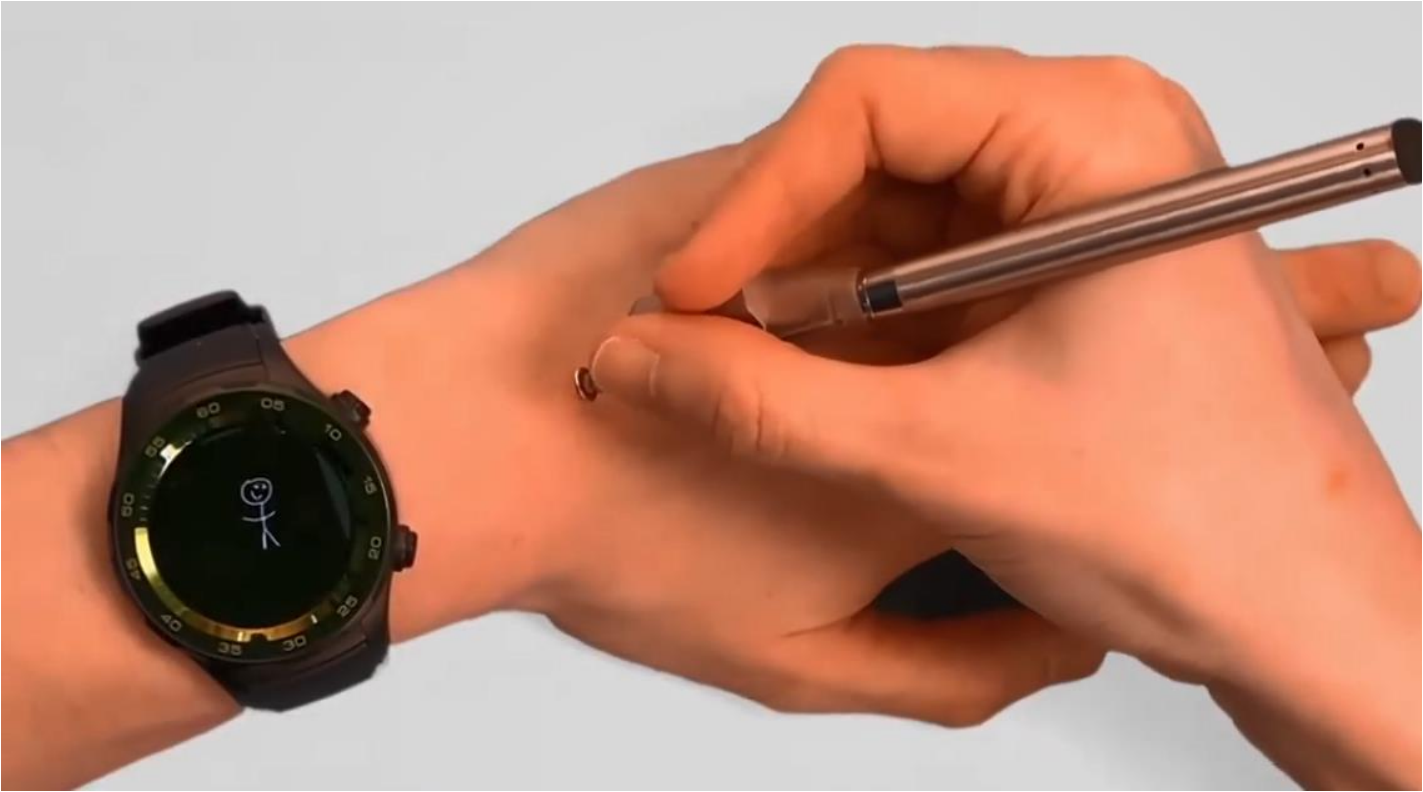}}
		
	\subfigure[A walking cycle sequence. Left: input hand-drawn sketch, middle: inflated 3D model with control points, right: walking cycle animation created by recording trajectories of individual control points specified by the user~\cite{Dvoroznak20-SA}.]{
		\label{fig:monster-mash}
		\includegraphics[width=0.8\columnwidth]{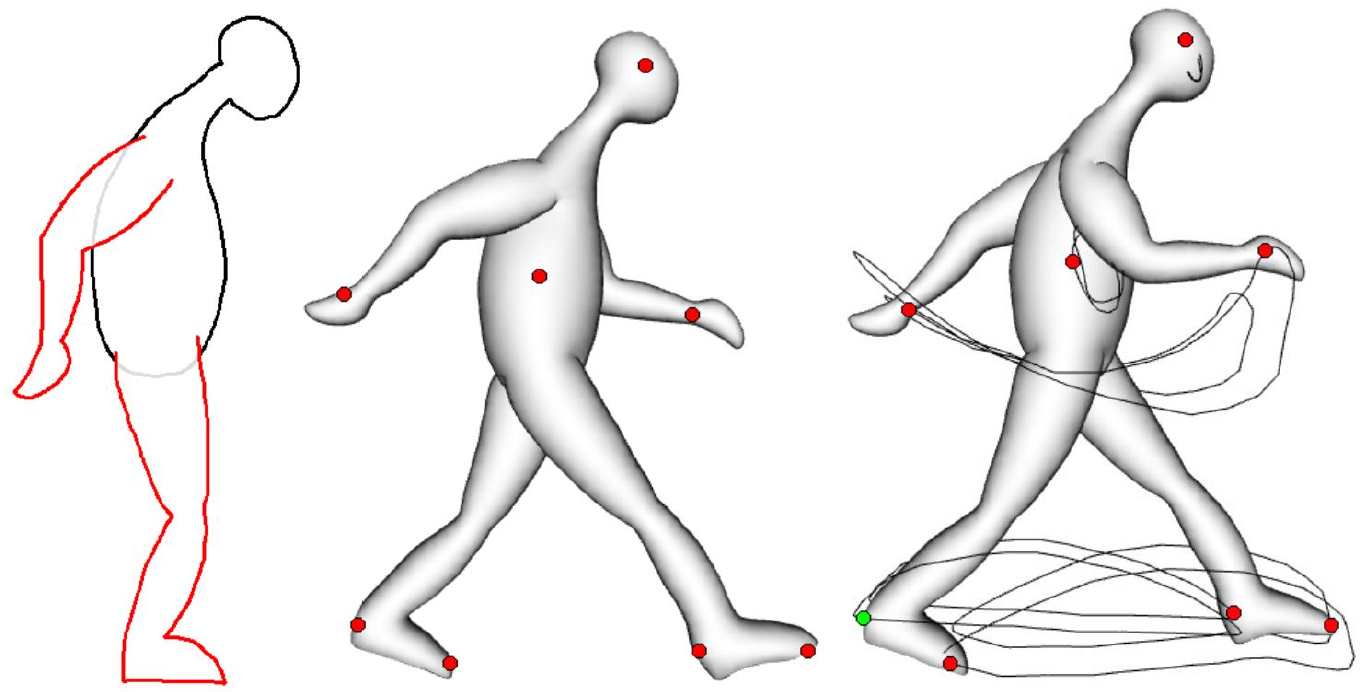}}
	\caption{  Novel applications that sketch supports.}
	\label{fig:killer-apps}
\end{figure}

Sketch also provides some unique opportunities compared to photos: (i) As a counterpoint to the sparsity challenge, sketch often lacks distracting background clutter compared to photos, which can benefit automated analysis \cite{yu2017sketch}. (ii) If captured appropriately, the sequential nature of sketch generation can further be exploited to benefit analysis compared to static images \cite{sarvadevabhatla2020pictionary}. (iii) The sparse and sequential nature of sketch also provides opportunities for high quality sketch generation, where image generation is hampered by the need to fill in pixel detail \cite{ha2018sketchrnn,song2018learning,das2021cloud2curve}. (iv) Sketches can serve as a computer-interaction modality in a way that photos cannot \cite{Yu_2016_CVPR,sarvadevabhatla2020pictionary}, due to the intuitive way humans can generate them without training. 
{
For example, {people} without professional painting training can do casual sketch-based design via sketch-to-photo generation techniques, \eg, scene photo generation~\cite{Gao2020SketchyCOCO}.
Figure~\ref{fig:watch-my-painting} presents another example: People could sketch on the back of the hand to make notes conveniently, and the sketch could be shown and recorded in the watch~\cite{Schrapel2020WatchMP}.
(v) Sketches natively can express motion trajectories, thus can be applied to dynamic modeling. As shown in Figure~\ref{fig:monster-mash}, walking cycle animation can be created by recording trajectories of individual control points specified by the user~\cite{Dvoroznak20-SA}.
}

Given these unique challenges and opportunities, it is often beneficial to design sketch-specific models to obtain best performance in various sketch-related applications.

\begin{figure*}[!t]
\begin{center}
\includegraphics[width=\textwidth]{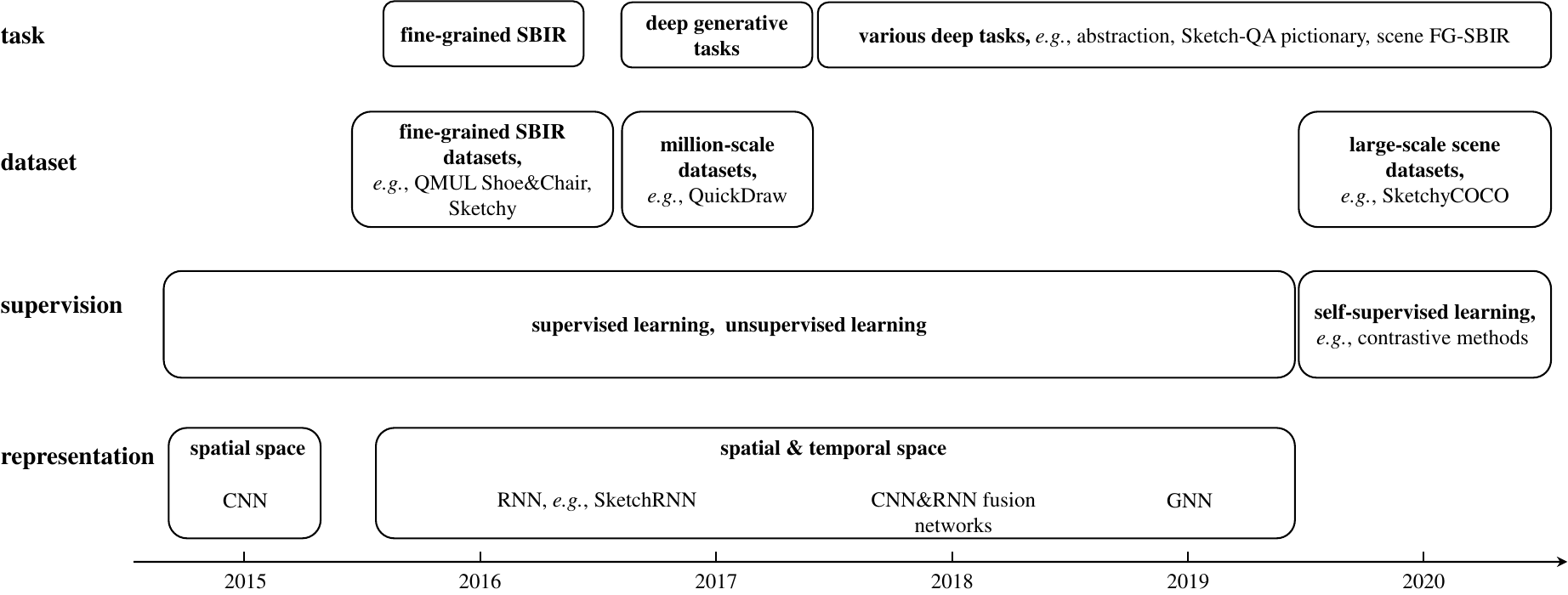}
\end{center}
   \caption{Milestones of deep learning based free-hand sketch research, from the perspectives of task, dataset, supervision, and representation. {Note that self-supervised learning is a branch of unsupervised learning.    }
}
\label{fig:milestones}
\end{figure*}

\subsection{A Brief History of Sketch in the Deep Learning Era}

In the past five years, the free-hand sketch community has developed rapidly as summarized by Figure~\ref{fig:milestones} from the perspectives of: tasks, datasets, representations and supervision. 
(i) In 2015, Sketch-a-Net~\cite{yusketch2015BMVC} was proposed as a CNN engineered specifically for free-hand sketch. It gained note as the first to achieve a recognition rate surpassing humans and helped to popularize deep learning for sketch analysis. 
(ii) In 2016, three fine-grained\footnote{{The phrase ``fine-grained'' in this paper has different meanings according to the context. For sketch tasks, fine-grained sketch-based image retrieval means instance-level sketch-photo matching, while other fine-grained tasks (\eg, generation, segmentation) emphasize that machine needs to perceive sketches on stroke or part or group levels. For sketch datasets, ``fine-grained'' datasets mean that their sketches provide visual details and/or detailed manual annotations (\eg, stroke/part/group level annotations, instance-level pairing information).}} sketch-based image retrieval (FG-SBIR) datasets were released, \ie, QMUL Shoe~\cite{Yu_2016_CVPR} and Chair~\cite{Yu_2016_CVPR}, and Sketchy~\cite{sangkloy2016sketchy}. 
Combined with deep triplet ranking~\cite{schroff2015facenet}, these fine-grained cross-modal datasets motivated a wave of follow-up FG-SBIR and other fine-grained tasks. 
(iii) In 2017, Google released a million-scale sketch dataset, \ie, Google QuickDraw, via the online game ``QuickDraw''. QuickDraw contains over 50M sketches collected  from players around the world, making it a rich and diverse dataset.
Furthermore, based on the QuickDraw dataset, Ha \etal proposed ``SketchRNN'', a RNN-based deep Variational Auto Encoder (VAE) that can generate diverse sketches \cite{ha2018sketchrnn}. This work motivated the community to go beyond considering sketches as static pictures to be processed by CNN; and inspired subsequent work to use stroke sequences as input and study temporal processing of sketches. 
In 2017, 
some sketch-based deep generative image models~\cite{Sangkloy_2017_CVPR} began to appear in the top conferences in computer vision.
(iv) From 2018 to date, based on deep learning techniques, various novel methodologies -- \eg,  sketch hashing~\cite{xu2018sketchmate}, sketch transformers \cite{xu2019multi}; and applications -- \eg,  sketch abstraction~\cite{muhammad2018learning},  sketch-based photo classifier generation~\cite{Hu_2018_CVPR}, sketch perceptual grouping~\cite{li2018universal}, and sketch vectorization \cite{das2021cloud2curve} have been proposed. 
See Figure~\ref{fig:milestones} for a chronological summary.

%

\begin{table*}[!t]

\small
\caption{Summary of the representative sketch datasets. 
Both `grouping'' and ``segmentation'' annotations refer to stroke-level. ``K'' and ``M'' mean ``thousand'' and ``million'', respectively. ``Cat.'' means ``category''. Stroke ``\checkmark'' denotes sketches provided as SVG files or coordinate arrays.}
\label{table:dataset_table}
\begin{center}
\resizebox{ \textwidth}{!}{
\begin{tabular}{ l | c | c | l | c | c | c | c | l | l }
\hline
 \tabincell{c}{Single-Modal \\ Datasets} & \tabincell{c}{Fine- \\ Grained} & Public & Modalities \& Sample Amount & Cat. &  Stroke   & \tabincell{c}{Object/ \\ Scene} & \tabincell{c}{Instance \\ Pairing} & {Annotations} & Remarks\\
\hline

TU-Berlin~\cite{eitz2012humans} &  & \checkmark & 20K sketches & 250 & \checkmark  & o  &  -  & {class} &   \\

QuickDraw~\cite{ha2018sketchrnn} &  & \checkmark & 50M+ sketches & 345  & \checkmark  & o  & -  & {class}  &   \\

QuickDraw-5-step~\cite{choi2019sketchhelper} &  &  & 38M+ sketches & 345  &   & o  & -  & {class}  &   \\

SPG~\cite{li2018universal} & \checkmark & \checkmark  & 20K sketches & 25 & \checkmark  & o  & - & {class, grouping}  &   \\

SketchSeg-150K~\cite{qi2019sketchsegnet+} & \checkmark &    & 150K sketches  & 20 & \checkmark  & o  & - & {class, segmentation}  & 57 semantic labels  \\

SketchSeg-10K~\cite{wang2019spfusionnet} & \checkmark & \checkmark & 10K sketches  & 10 &   & o  & - & {class, segmentation}  &  24 semantic labels \\

SketchFix-160~\cite{sarvadevabhatla2017object}  &  & \checkmark  &  3904 sketches  & 160  & \checkmark  &  o & -  & {class, eye fixation}  &   \\

{Sheep 10K}~\cite{aaron_sheep}  & \checkmark & \checkmark  &  10K sheep sketches  & 1  & \checkmark  &  o & -  & {class}  &   \\

{COAD}~\cite{tirkaz2012sketched}  & \checkmark & \checkmark  &  620 sketches  & 20  & \checkmark  &  o & -  & {class}  &   \\

\hline

&    &    &   &   &   &   &   &   &   \\
 \tabincell{c}{Multi-Modal \\ Datasets} & \tabincell{c}{Fine- \\ Grained} & Public & Modalities \& Sample Amount & Cat. &  Stroke  & \tabincell{c}{Object/ \\ Scene} & \tabincell{c}{Instance \\ Pairing} & {Annotations} & Remarks\\

\hline
&    &    &   &   &   &   &   &   &   \\
QMUL Shoe~\cite{Yu_2016_CVPR}    & \checkmark & \checkmark  & 419 sketches, 419 photos  &  1 &  & o  & \checkmark  & {pairing,triplet,attribute}  &  21 binary attributes  \\

QMUL Chair~\cite{Yu_2016_CVPR}   & \checkmark  & \checkmark  & 297 sketches, 297 photos  &  1 &   & o  &  \checkmark &  {pairing,triplet,attribute} &   15 binary attributes \\

QMUL Handbag~\cite{Song_2017_ICCV}  & \checkmark  &  \checkmark & 568 sketches, 568 photos  & 1  &   & o  & \checkmark  & {pairing}  &   \\

Sketchy~\cite{sangkloy2016sketchy}   & \checkmark  & \checkmark  & 75K sketches, 12K photos  & 125  &  \checkmark & o  &   &  {class} &  12K objects \\

Sketch\&UI~\cite{huang2019swire}   & \checkmark  &   & 1998 sketches, 1998 photos & 23  &   &  o & \checkmark  &  {class, pairing} & UI  \\

QuickDrawExtended~\cite{dey2019doodle} &  & \checkmark & 330K sketches, 204K photos & 110   &   & o  &   &  {class} &   \\

SketchTransfer~\cite{{lamb2020sketchtransfer}} &  &  & \tabincell{l}{112.5K sketches,\\ 90K CIFAR-10 photos} & 9   &   & o  &   &  {class} & resolution of 32x32  \\

TU-Berlin Extended~\cite{Zhang_2016_CVPR}  &   &    & 20K sketches, 191K photos  & 250  &   & o  &   &  {class} &   \\

Sketch Flickr15K~\cite{hu2013performance} &   &  \checkmark  & 330 sketches, 15K photos  & 33 &  & o  &   & {class}  &   \\

Aerial-SI~\cite{jiang2017sketch,jiang2017retrieving} &   &    &  400 sketches, 3.3K photos & 10  &   & o, s &   & {class}  & aerial scene   \\

HUST-SI~\cite{wang2016deep} &   &  \checkmark  & 20K sketches, 31K photos  & 250  & \checkmark  & o  &   & {class}  &   \\

SBSR~\cite{eitz2012sketch} &  &  \checkmark  & 1814 sketches, 1814 3D models & 161  &   & o  &   & {class}  &   \\

SHREC'13~\cite{li2014comparison} & \checkmark &  \checkmark  & 7200 sketches, 1258 3D models & 90  &   &  o &   & {class}     &   \\

SHREC'14~\cite{li2014shrec} & \checkmark &  \checkmark  & 12680 sketches, 8987 3D models & 171  &   & o  &   & {class} &   \\

PACS DG~\cite{Li_2017_ICCV} &  &  \checkmark  & \tabincell{l}{ 9991 (sketches, photos, \\ cartoons, paintings)}  &  7 &   & o  &   &  {class} & \tabincell{c}{domain \\ generalization}   \\

Flickr1M~\cite{xiao2015sketch} &  &   & 500 sketches, 1.3M photos  &  100 &   &  o &   & {class}  &   \\

Cross-Modal Places~\cite{castrejon2016learning} &  &  \checkmark  & \tabincell{l}{16K sketches, 11K descriptions,\\ 458K spatial texts, 12K clip arts, \\ 1.5M photos}  & 205  &   & s  &   & {class} &   \\

{SketchyScene}~\cite{zou2018sketchyscene}  & {\checkmark} & {\checkmark}  &  {29K sketches, 7K photos}  &   &   &  {s} &  {\checkmark}  & {pairing, segmentation}  &   \\

DomainNet~\cite{peng2019moment} &  &  \checkmark  & \tabincell{l}{ 0.6M (cliparts, infographs, \\paintings, 	QuickDraw skteches, \\ real photos,\\ professional pencil sketches)}  & 345  & \checkmark  & o &   & {class} &   \\

{SketchyCOCO}~\cite{Gao2020SketchyCOCO} &  \checkmark &  \checkmark  & \tabincell{l}{ 14K+ (sketches, photos, \\ edge-maps)}  & 17  & \checkmark  & o, s &  \checkmark  & {\tabincell{l}{class, pairing, \\ five-tuple, \\ segmentation}} &  \tabincell{l}{3 background classes  \\  14
foreground classes} \\

{SceneSketcher}~\cite{liu2020scenesketcher} & \checkmark &  \checkmark  & \tabincell{l}{  1225 scene \\ sketch-photo pairs}  & 14  & \checkmark  & o, s & \checkmark  & {\tabincell{l}{class,pairing   \\   segmentation}} &   \\
\hline

\end{tabular}
}
\end{center}

\end{table*}


\section{Free-Hand Sketch Datasets}
\label{sec:datasets}

In the last decade numerous new free-hand sketch datasets have been collected, to satisfy the need for large-scale deep network training, and the growing diversity of sketch-related tasks considered by the community. This section will summarize these datasets, and further discuss some of the unique challenges in sketch-related data collection.



\subsection{Sketch Datasets}
Free-hand sketch datasets can be grouped in terms of: (i) single vs. multi-modal, and (ii) coarse vs. fine-grained. Single-modal datasets consist only of sketches and are typically used for recognition, sketch-sketch retrieval, grouping, segmentation, and generation. Multi-modal datasets support cross-modal tasks by pairing sketches  with samples from other modalities such as natural photo, 3D shape, text, or video. These are mainly used for cross-modal retrieval/matching, or cross-modal generation/synthesis. Coarse-grained datasets (\eg, TU-Berlin~\cite{eitz2012humans}, QuickDraw~\cite{ha2018sketchrnn}) are usually used for sketch recognition, sketch retrieval; while fine-grained datasets (\eg, QMUL Shoe~\cite{Yu_2016_CVPR}) provide fine-grained visual details and manual annotations. 

More specifically, coarse-grained single-modal datasets \cite{eitz2012sketch,ha2018sketchrnn} support sketch recognition  and retrieval; while coarse-grained multi-modal datasets (\eg, QuickDraw-Extended \cite{dey2019doodle}) support category-level sketch-based image retrieval.
Fine-grained single-modal datasets \cite{li2018universal,qi2019sketchsegnet+} support perceptual grouping, segmentation, and parsing.  
Fined-grained multi-modal  datasets (\eg, QMUL Shoe~\cite{Yu_2016_CVPR}) provide the instance-level pairing information to support retrieval. 

Table~\ref{table:dataset_table} summarizes representative sketch datasets of each type in terms of: modalities, size, number of categories, stroke information, annotation, \etc.
{Note that SVG files are able to generate static picture files such as JPEG and PNG, whilst these static files cannot store or provide the original drawing process (stroke ordering).} We exclude some well-known but overly-small datasets such as \cite{kato1992sketch}.

\subsection{{Unique Challenges of Sketch Collection}}
\label{sec:Challenge_of_Sketch_Collection}
\keypoint{Sketch Collection Strategies}
Existing collection approaches mainly include:
(i) bespoke creation by researchers \cite{li2018universal,wang2019spfusionnet,qi2019sketchsegnet+},
(ii) crowd-sourcing selecting and matching on existing datasets, \eg, Doodle2Sketch QuickDraw-Extended~\cite{dey2019doodle},
(iii) crowd-sourcing drawing from scratch, \eg, QMUL Shoe~\cite{Yu_2016_CVPR}, Sketchy~\cite{sangkloy2016sketchy}.
(iv) collecting via online drawing games, \eg, Google QuickDraw~\cite{ha2018sketchrnn}.
(v) Web crawling of existing sketches \cite{peng2019moment,Li_2017_ICCV}. 
In particular, for fine-grained multi-modal sketch datasets,  crowd-sourcing is widespread, since fine-grained drawing, selection, and matching are time-consuming.~{Note that a sketch dataset's potential applications are determined by both its collection and annotation protocol.}

\keypoint{Sketch Collection Challenges}
Free-hand sketch poses some unique data collection challenges compared to other image types:
(i) \textbf{Time-Sequence Nature} Sketching is a dynamic and temporally extended process. Thus, collecting sketches as static raster images  (\eg, JPEG, PNG) is very limiting, and recording vector representation (\eg, SVG\footnote{\url{https://en.wikipedia.org/wiki/Scalable_Vector_Graphics}}) together with stroke position and timing is preferred to support research. As a consequence, this means that collecting by web-crawling (which typically retrieves raster images), is less useful. 
(ii) \textbf{Cross-Modal Pairing} Collecting cross-modal datasets 
provides the additional challenge of pairing sketches and associated data in other modalities. One can start with existing images and sketches and pair them \cite{yi2014bmvc}, or draw sketches specifically corresponding to given examples in other modalities \cite{sangkloy2016sketchy,Yu_2016_CVPR}. But both options are time-consuming. 
(iii) \textbf{Demographic Information} Since sketches are human-created, rather than pixel-perfect images captured by camera, it is important to ensure that sketch datasets are created by diverse demographics to ensure they are representative (an even stronger version of the challenge \cite{de2019does} posed by photo images). Furthermore, meta-data about the artists (\eg, gender, nationality, skill level) may be important to store, both in order to study how different humans sketch, and also to ensure that any sketch-based applications are fair and unbiased \cite{barocas-hardt-narayanan}. However to our knowledge, this kind of meta-data has not been recorded in existing datasets.

\keypoint{{Discussion}} {These challenges all mean that the standard computer vision approach of web-crawling is poorly suited to collection of sketch data: It does not typically retrieve vector-graphic and time-stamped sketch generation, does not make it easy to obtain matched cross-modal pairs, and does not come with any demographic information. For these reasons, bespoke creation, crowd-sourcing, and generation as a byproduct through gamification are the recommended methods of collection.}

\begin{figure*}[!t]
\begin{center}
\includegraphics[width=\textwidth]{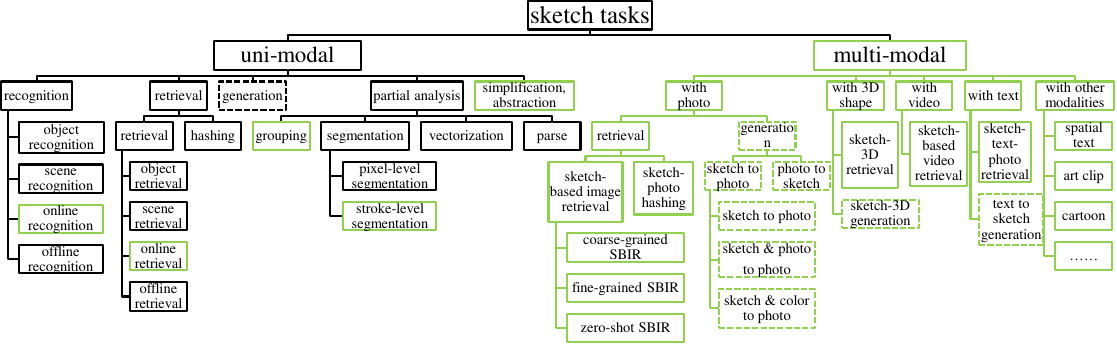}
\end{center}
   \caption{A tree diagram of the sketch task taxonomy. Generative tasks are framed by dashed lines. Sketch domain-unique tasks are framed by green lines. Best viewed in color.}
\label{fig:taxonomy_tree_map}
\end{figure*}

\section{Tasks and Methodology Taxonomy}
\label{sec:taxonomy}

{
In this section, we aim to provide an overview of deep learning related tasks and methods from the perspective of the whole free-hand sketch research area, rather than categorizing methods for a specific task.
We observe that several trends emerging in contemporary sketch research including: 
(i) More novel tasks are continually being proposed, each of which has different task-specific challenges, and thus different task-specific methods/designs. 
This inspires us to review the existing methods from a task-driven perspective.
(ii) Free-hand sketch is often associated with data from other modalities.
This inspires us to categorize the existing tasks on the basis of the data modalities involved. 
}

According to the data modalities involved, free-hand sketch related tasks can be divided into single- and multi-modal tasks, with single-modality sketch analysis techniques  often used as building blocks for multi-modal methods. 
This section will define the popular sketch analysis tasks and introduce the corresponding deep learning methods, providing a detailed taxonomy. Figure~\ref{fig:taxonomy_tree_map} provides a tree diagram of the existing free-hand sketch tasks. 
{
The main advantages of this task taxonomy include:
(i) Straightforward and efficient. The tree has balanced depth and width, and does not contain redundant nodes.
(ii) Extensible.  The single-modality and multi-modal sub-trees are uncoupled for independent update. E.g., the single-modality sub-tree can remain fixed when we insert new modalities in future.
}

{
Some uni-modal tasks listed in Figure~\ref{fig:taxonomy_tree_map} are also studied in the natural photo domain, while the sketch-based multi-modal tasks are unique.
}


\subsection{Uni-Modal Tasks: Pure Sketch Analysis}
These tasks study sketches in isolation without other data modalities. Key deep learning-based applications in this area include 
 recognition, retrieval/hashing, generation, grouping, segmentation, and  abstraction.

\subsubsection{Recognition}
Sketch recognition~\cite{eitz2012humans} aims to predict the class label of a given sketch, which is one of the most fundamental tasks in computer vision. It has a variety of  practical applications including: {interactive drawing systems that provide feedback to users~\cite{matsui2016drawfromdrawings}, sketch-based science education ~\cite{forbus2018sketch}, games \cite{sarvadevabhatla2020pictionary,ha2018sketchrnn}, \etc.} Both object~\cite{eitz2012humans,ha2018sketchrnn} and scene~\cite{ye2016human,xie2019deep} categories have been studied from a recognition perspective. Notably sketch recognition techniques underpin the popular web game QuickDraw and WeChat mini-app Caihua Xiaoge, both released by Google.

Sketch recognition can be categorized into (i) offline and (ii) online recognition settings. Offline recognition systems take the whole sketch as input and predict a class label based on the complete sketch. Online recognition systems take
the accumulated sketch strokes and continuously predict the class label, during sketching. Offline recognition methods are more common, but online methods can be used in more interactive real-time applications such as real-time drawing guidance~\cite{choi2019sketchhelper}, tracing, and interactive sketch retrieval.


There are several current trends in sketch recognition, building on  underpinning deep learning progress: 
(i) From raster image  to sequence representation;
(ii) From global representation to local analysis;
(iii) From Euclidean (CNN, RNN based) to topological analysis (GNN based),
(iv) From fully-supervised learning to self-supervised learning. 

Numerous deep models have now been proposed for free-hand sketch recognition~\cite{seddati2015deepsketch,zhang2016deep,guo2016building,Zhang_2016_CVPR,ballester2016performance,seddati2016deepsketch,zhang2019learning,seddati2016deepsketch2image,zhang2019cousin,zhang2020hybrid,jain2020transsketchnet,jiao2020tactile,jia2020coupling}.
In the following, we review these models from the perspectives of
network architectures and loss functions. Data augmentations will be discussed separately in Section~\ref{sec:data_augmentations}.


\begin{table*}[!t]

\scriptsize

\caption{Comparison of representative sketch recognition networks. ``--'' indicates not mentioned or unclear in the original paper. Reported performance is top-1 accuracy. Abbreviations in this table:  ``stroke accu. pic.'': stroke accumulated pictures; ``R-FC'': residual fully-connected layer; \cut{``cat.'': category;} ``pad.'': padding; ``tru.'': truncation; ``augm.'': specific augmentations; ``tran.'': transformer.}
\label{table:networks_for_sketch_recognition}
\begin{center}
\resizebox{\textwidth}{!}{
\begin{tabular}{ c | c | c | c | c | c | c | c | c | c | c}
\hline
Year & Model & Architecture & Layers & Params & Ensemble & Pretrain & Input & Preprocess & Dataset & Accuracy \\
\hline
2015 & Sketch-a-Net~\cite{yusketch2015BMVC} & CNN & 5 conv. & 8.5M& \checkmark & & picture & augm.~\cite{yusketch2015BMVC} & TU-Berlin~\cite{eitz2012humans} 250 cat. & 0.7490  \\
2016 & AlexNet-FC-GRU~\cite{sarvadevabhatla2016enabling}  & \tabincell{c}{CNN-to-RNN \\ cascaded}  & --  & -- &   & & \tabincell{c}{stroke \\ accu. pic.} & --  & TU-Berlin 160 cat. & 0.8510 \\
2018  & SketchMate~\cite{xu2018sketchmate,xu2020learning
} & RNN  & 2 GRU  & -- &   &  & stroke vector & tru. \& pad.~\cite{xu2018sketchmate} & QuickDraw 3.8M~\cite{xu2018sketchmate} & 0.7788 \\
2018   &  SketchMate~\cite{xu2018sketchmate,xu2020learning
} & \tabincell{c}{CNN-RNN \\ dual-branch}  & \tabincell{c}{5 conv. \\ \& 2 GRU}  & -- &   &  & \tabincell{c}{picture \\ \& stroke vector} & tru. \& pad.~\cite{xu2018sketchmate} & QuickDraw 3.8M~\cite{xu2018sketchmate}   & 0.7949 \\
2017 & Jia \etal~\cite{jia2017sequential}  & \tabincell{c}{RNN-RNN\\ dual branch} &  -- & -- &   \checkmark & \checkmark & \tabincell{c}{CNN features \\ of stroke \\ accu. pic.} & \tabincell{c}{reflection, \\ rotation, \etc.} & TU-Berlin & 0.9220 \\
2017 &  DVSF~\cite{he2017sketch} & \tabincell{c}{R-FC and RNN \\ dual branch}  & --  & -- &   & \checkmark & \tabincell{c}{CNN features \\ of stroke \\ accu. pic.} & -- & TU-Berlin  & 0.7960 \\
2018 & FBin DAB-Net~\cite{prabhu2018distribution}  & binary CNN  & --  & -- &   &  & picture & --  & TU-Berlin & 0.7370 \\
2018  & RNN$\rightarrow$CNN~\cite{li2018sketch}  &  \tabincell{c}{RNN-to-CNN \\ cascaded} & \tabincell{c}{2 LSTM \\ \& 5 conv.}  & -- &   & \checkmark & stroke vector & augm.~\cite{yu2017sketch}  & TU-Berlin & 0.7849 \\

2019 & multi-graph tran.~\cite{xu2019multi}  & GNN  &  4 tran. & 10M &   &  & stroke vector & -- & QuickDraw subset~\cite{xu2019multi} & 0.7070 \\

\hline
\end{tabular}
}
\end{center}

\end{table*}

\keypoint{Networks}
Figure~\ref{fig:milestones} (below) summarizes the evolution of the deep learning-based sketch representations.
Moreover, Table~\ref{table:networks_for_sketch_recognition} lists  various networks that are engineered for free-hand sketches and capable of sketch recognition. We next introduce  some representative networks.

(i) Sketch-a-Net~\cite{yusketch2015BMVC,yu2017sketch} was the first deep CNN designed for free-hand sketch. 
Compared with classic photo-oriented CNN architecture~\cite{krizhevsky2012imagenet}, the sketch-specific aspects of its architecture mainly include:
(a) Considered the sparse low-texture nature of sketch, larger size ($15 \times 15$) first layer filters are used to capture more context. 
(b) Local response normalization~(LRN)~\cite{krizhevsky2012imagenet} layers are removed for faster learning without sacrificing performance, since LRN is for ``pixel brightness normalization'', but most sketches are binary images. (c) Two novel sketch-specific data augmentation strategies are proposed, leveraging stroke appearance and stroke sequence. (d) Finally, an ensemble of Sketch-A-Net were combined using joint Bayesian fusion~\cite{chen2012bayesian}.

(ii) 
Sarvadevabhatla \etal~\cite{sarvadevabhatla2016enabling}  proposed a sketch recognition network to leverage the sequential process of sketching, where each training sketch is plotted as a continuous sequence of cumulative stroke pictures and the corresponding AlexNet~\cite{krizhevsky2012imagenet} based deep features will be sent into a Gated Recurrent Unit (GRU)~\cite{cho2014learning} network in sequence.
This network is also able to work in online recognition mode, since it involves the intermediate status of the sketch.
{Furthermore, Jia \etal~\cite{jia2017sequential} proposed a multiple feature based model to improve this idea and obtained good performance on sketch recognition as reported in Table~\ref{table:networks_for_sketch_recognition}, where multiple GRU networks separately encode multiple features of the cumulative stroke groups and their outputs are combined by time-step-based weights.}

(iii) Similarly, He \etal \cite{he2017sketch} proposed the deep visual-sequential fusion (DVSF) net to capture spatial and temporal 
patterns of sketches simultaneously. For each training sketch, its three accumulation sub-pictures (with $60\%$, $80\%$, $100\%$ of strokes) go through three-way CNNs (ResNet-18~\cite{he2016deep}) to produce deep features, which are  fed into both visual and sequential networks. In particular, the visual and sequential networks are implemented by residual fully-connected (R-FC) and Residual Long Short Term Memory~(R-LSTM)~\cite{oord2016pixel} layers respectively.
The visual and sequential paths are integrated by a fusion layer for final recognition.

(iv) 
In 2017, Ha and Eck proposed the groundbreaking  SketchRNN~\cite{ha2018sketchrnn}, which performs representation learning through its Variational Inference (VI)~\cite{kingma2013auto} based sequential sketch generation model. 
Distinctively different to the prior stroke accumulated sub-picture representations, the key points of sketch strokes are directly fed into the RNN backbone of SketchRNN.
In particular, as illustrated in Figure~\ref{fig:motivations}, 
each key point is denoted as a vector consisting of two coordinate bits (\ie, horizontal and vertical coordinates) and the corresponding flag bits. The flag bits indicate the start/end of a stroke by pen state.
{Although initially proposed for generative modeling, the encoder backbone of SketchRNN also performs well for sketch recognition\footnote{\url{https://github.com/payalbajaj/sketch_rnn_classification}}.}

(v) 
Xu \etal proposed the sketch hashing network  SketchMate~\cite{xu2018sketchmate}, where the backbone is a CNN-RNN dual branch network, using CNN to extract abstract visual concepts and RNN to model human  temporal stroke order.
The CNN branch takes in the raster pixel sketch pictures; and the RNN branch process the vector sketch (\ie, key point coordinates). The branches are combined by a late-fusion layer.
This network demonstrates the complementarity of visual and temporal embedding spaces of sketch representation.
This CNN-RNN dual-branch modeling idea has been widely applied to other sketch tasks, \eg, SPFusionNet~\cite{wang2019spfusionnet} for sketch semantic segmentation.
In addition to the parallel pipelines of CNN and RNN, some cascaded pipelines~(\eg, RNN-to-CNN~\cite{li2018sketch})  have also been studied.

(vi) A sketch can be represented as the sparsely connected graphs in topological space. Multi-Graph Transformer~(MGT) \cite{xu2019multi} is a GNN model that learns both geometric structure and temporal information from sketch graphs. MGT injects domain knowledge into Graph Transformers\footnote{{Transformer is essentially a GNN that encodes input as a fully-connected graph.}} through sketch-specific graphs. In particular, MGT represents each sketch as multiple intra-stroke and extra-stroke graphs, to model its local and global topological stroke structure, respectively.

(vii) While prior approaches to sketch recognition are based on supervised learning (\eg, \cite{yu2017sketch,sarvadevabhatla2016enabling,he2017sketch}), \cite{xu2020deep} provided the first investigation of  self-supervised representation learning for sketch, 
{proposing a rotation- and deformation- based deep self-supervised model (rot. \& def. model). This model uses multi-branch CNN and TCN network to represent sketch in a self-supervised setting.}



\begin{figure}[!t]
\begin{center}
\includegraphics[width=0.7\columnwidth]{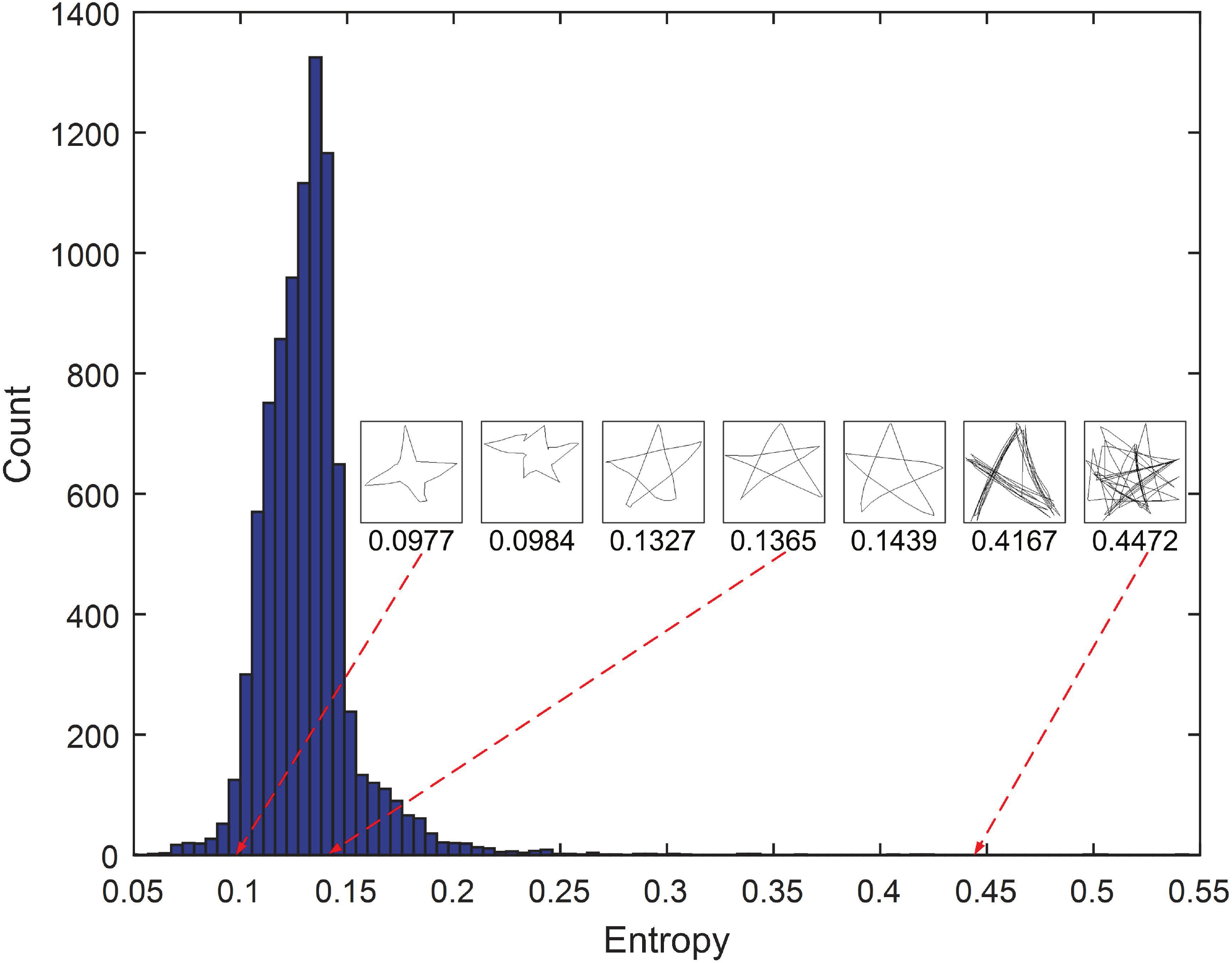}
\end{center}
   \caption{ \textcolor{black}{Image entropy histogram of  9K sketch `stars'~\cite{xu2018sketchmate}.
   The blue bars denote the bin counts within different entropy ranges.
    Some representative sketches corresponding to different entropy values are illustrated.}}
\label{fig:star}
\end{figure}

\keypoint{Loss Functions} 
Most of the previous deep sketch recognition methods use cross-entropy softmax loss to train deep neural networks. An active research question is whether sketch-specific loss functions can help to further improve recognition performance. 
To that end, Xu \etal proposed the sketch-specific center loss
~\cite{xu2018sketchmate} for million-scale sketches, based on a staged-training strategy. 
The basis is that the image entropy distribution of each sketch category is a truncated Gaussian distribution {(see Figure~\ref{fig:star} for an example)}.
 Inspired by classic Bayesian decision theory~\cite{duda2012pattern}, Mishra \etal proposed a novel metric loss to drive the pretrained deep neural network to minimize the Bayesian risk of mis-classifying sketch pairs that were randomly selected within each mini-batch~\cite{mishra2018deep}.
Based on this Bayesian risk loss, sketch recognition needs two-stage training. After obtaining the features, a linear SVM~\cite{cortes1995support} is trained as the classifier.

\cut{
\keypoint{{Comments}}
{
{Both these two losses have a common disadvantage that fails in end-to-end training} {that refers to training a machine learning system as a differentiable whole to learn solutions from data space directly}.
}}

\keypoint{{Summary and Discussion}} {In this subsection, we reviewed sketch recognition related deep learning works, from the perspective of architecture and loss functions.
Promising areas of future work include: Online sketch recognition, motivated by practical human-computer interaction applications; going beyond the mainstream line of supervised sketch recognition to investigate semi-supervised, self-supervised \cite{bhunia2021vectorization} and unsupervised learning for recognition; zero-shot~\cite{xie2019deep} and few-shot sketch recognition~\cite{pan2020teach}; and multi-task learning \cite{seddati2016deepsketch2image, Liu_2019_CVPR} to simultaneously solve sketch recognition with other tasks.}

\subsubsection{Retrieval and Hashing}

Sketch retrieval~\cite{wang2015spatial, creswell2016adversarial,choi2019sketchhelper} methods aim to use a query sketch to retrieve similar samples from a gallery or database of sketches. {Sketch retrieval is a challenging task due to abstraction, intra-class variation, drawing style variation, and feature sparsity. These properties make it difficult to localize repeatable feature points across sketches (\eg, the manner of SIFT~\cite{lowe2004distinctive}) in order to perform classic interest-point based retrieval approaches~\cite{lowe2004distinctive}.} In the deep learning era, end-to-end feature learning has outperformed shallow features on various retrieval tasks in computer vision, and CNNs have also been used for sketch retrieval.

Common practice in image retrieval is to use CNNs to learn a vector embedding, and then perform retrieval/matching as Nearest Neighbor search. Most existing deep sketch retrieval models work in a similar metric learning manner, with research focusing on CNN architectures and loss function designs for effective sketch matching. Wang \etal \cite{wang2015spatial} proposed a representative sketch retrieval pipeline, which has two key components: A pure convolutional layer based Siamese CNN backbone, and an $\ell_{1}$ norm distance based pair-wise loss between query and gallery images. 
The idea is two-fold: 
(i) Use the convolutional feature map to preserve the spatial information for sketches without point correspondence.
(ii) Compute distance in feature space, and optimize for similar pairs to be nearby while different pairs to be far apart. 

Given the growing number of images available, there is also an increasing concern about scalability of retrieval, leading to studies of hashing-based methods where all sketches are encoded and searched as binary hash code vectors, rather than real-valued vectors. Xu \etal \cite{xu2018sketchmate} proposed the first deep sketch-hashing model. Their deep sketch hashing used a dual-branch CNN-RNN network to exploit both global appearance and local sequential stroke information, as well as a new center loss variant to ensure the learned embedding is more semantically meaningful. 

The sketch retrieval/hashing methods mentioned so far exploit supervised information. If class labels are unavailable, adversarial training can be used to learn a feature representation for sketches. 
Based on the Generative Adversarial Network (GAN)~\cite{goodfellow2014generative}, Creswell \etal \cite{creswell2016adversarial} proposed Sketch-GAN for unsupervised sketch retrieval, where both the query and gallery sketches are represented by the output features of the discriminator network.

\subsubsection{Generation}

\begin{table*}[!t]
\caption{{ Comparison of the representative sketch generation deep models.
``gen.'', ``rec.'', and ``com.'' denote ``generation'', ``reconstruction'', and ``completion'', respectively.} {``RNN+VAE'' means ``RNN backbone based VAE''.}}
\label{table:sketch-generation-frameworks}
\begin{center}
\resizebox{\textwidth}{!}{
\begin{tabular}{ l | c | c | l}
\hline
{Pipelines} & Representative Ref. & Applications & Advantages \& Disadvantages \\
\hline
{RNN+VAE} & SketchRNN~\cite{ha2018sketchrnn} &  gen., rec., com. & \tabincell{l}{A: brief, flexible. \\    D: scribble effect, single-class gen.} \\
{RNN+GAN} & SkeGAN~\cite{balasubramanian2019teaching} & gen., com. & \tabincell{l}{A: less scribble effect, faster convergence. \\   D: single-class gen.}  \\
{RNN+VAE+GAN} & VASkeGAN~\cite{balasubramanian2019teaching} &  gen., rec., com. & \tabincell{l}{ A: less scribble effect. \\     D: high complexity, single-class gen., slower convergence}  \\
BERT & \tabincell{l}{Sketchformer~\cite{ribeiro2020sketchformer} \\ Sketch-BERT~\cite{Lin2020Sketch-BERT}  } &  gen., rec., com.  & \tabincell{l}{A: good at handling longer stroke sequences  \\    D: more parameters, high complexity}  \\
{CNN+RL} & Doodle-SDQ~\cite{zhou2018doodle} & imitating a reference  & \tabincell{l}{A: can handle unseen classes.   \\     D: hybrid training (supervised learning \& RL), high complexity}  \\
{VAE+Renderer} & Cloud2Curve~\cite{das2021cloud2curve} & gen. rec.  & \tabincell{l}{A: Scaleable vector sketch generation. Long sketches. \\  D: High complexity. } \\
\hline
\end{tabular}
}
\end{center}

\end{table*}

\begin{figure*}[!t]
\begin{center}
\includegraphics[width=0.9\textwidth]{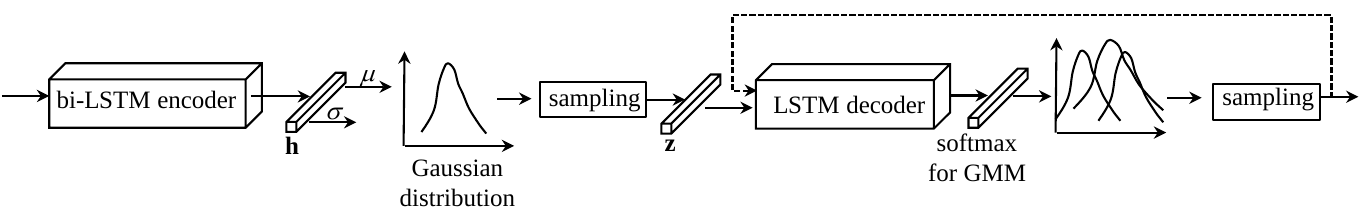}
\end{center}
   \caption{The pipeline of SketchRNN \cite{ha2018sketchrnn}. The {dashed} arrow line denotes the recurrent processing of LSTM decoder. For simplicity, the recurrent processing of bi-LSTM encoder is not shown here.}
\label{fig:sketch_RNN}
\end{figure*}

Sketch generation~\cite{ha2018sketchrnn,jaques2018learning,sasaki2019,li2020sketchman,ge2021creative,das2020beziersketch,das2021cloud2curve,bhunia2020pixelor} has grown rapidly in recent years as deep learning-based approaches easily outperform earlier classic sketch generators \cite{hinton2005motorProgram,li2017free}.  Sketch generation has several practical applications,  \eg, synthesizing novel pictures, assisting artist design, and finishing incomplete sketches and games \cite{bhunia2020pixelor}. It can be addressed using various deep learning tools, \eg, 
VAE~\cite{ha2018sketchrnn,jaques2018learning,jenal2019rnn,bhunia2020pixelor,das2020beziersketch}, GAN~\cite{balasubramanian2019teaching}, VAE-GAN~\cite{balasubramanian2019teaching}, {Bidirectional Encoder Representations from Transformers (BERT)~\cite{devlin2018bert}}, and reinforcement learning (RL)~\cite{zhou2018doodle,balasubramanian2019teaching}. 
{
We compare these pipelines and their representative models in Table~\ref{table:sketch-generation-frameworks}.   Most of these pipelines are flexible and able to use GRU, LSTM, Transformer as backbones to achieve the stroke-by-stroke generation.
}

{The seminal model SketchRNN~\cite{ha2018sketchrnn} is a sequence-to-sequence
VAE for conditional and unconditional generation of vector sketches. 
Its encoder and decoder are implemented by bidirectional RNN~\cite{schuster1997bidirectional} and unidirectional RNN, respectively. As stated earlier, free-hand sketches can be represented as a sequence of keypoints defining strokes. The main idea of SketchRNN is to simulate human sketching by sequential generation of these key points in terms of location and pen up/down status. }

As shown in Figure~\ref{fig:sketch_RNN}, 
the VAE encoder of SketchRNN takes vector sketches as input, and encodes it as a  vector ${\bf h}$, which is the  RNN's last hidden state. 
This  vector will be further encoded as two parameters ${\bf \mu}$ and ${\bf \sigma}$ to model a Gaussian distribution $N({\bf \mu}, {\bf \sigma})$,
from which a latent vector ${\bf z}$ will be sampled.
Then, the LSTM based VAE decoder will generate the  coordinates and pen states of the key stroke points, conditional on ${\bf z}$.
In particular, the coordinate and state for each key point is sampled from a Gaussian mixture model~(GMM), and also used as input for the next decoder step. To improve SketchRNN to deal with multi-class generation, Cao \etal \cite{cao2019ai} propose a generative model named as ``AI-Sketcher'', which is also a VAE based network.

Another line of work within sketch generation uses differentiable rendering \cite{zheng2018strokenet,das2020beziersketch,das2021cloud2curve} or reinforcement learning \cite{ganin2018imagePrograms,balasubramanian2019teaching,zhou2018doodle,mellor2019unsupervised,bhunia2020pixelor} to train policies that draw sketches iteratively according to different criteria such as adversarial training against human sketches \cite{ganin2018imagePrograms}. This line of work often considers factors not addressed by SketchRNN such as brush style and color. By considering an interpretable latent representation of sketches, such methods can also potentially be used to de-render sketches into programs or symbols \cite{ellis2017graphicProgram,egiazarian2020deep}.

Going forward, there are several emerging trends in sketch generation, notably:
(i) Fine-grained sketch generation~\cite{jenal2019rnn}. 
(ii) A novel evaluation metric ``Ske-score''~\cite{balasubramanian2019teaching}, aims to provide a better metric to quantify the goodness of generated vector sketches.
(iii) Transformer-based architectures ~\cite{ribeiro2020sketchformer,wieluch2020strokecoder} are being applied to sketch generation. (iv) Competitive generation, which aims to render understandable sketches in the fewest possible strokes \cite{bhunia2020pixelor}. (v) Finally there is scaleable vector-graphic generation \cite{das2020beziersketch,das2021cloud2curve}, which aims to generate sketch strokes via parametric strokes rather than standard waypoint lists.

\subsubsection{Grouping, Segmentation, and Parsing}

Compared with sketch recognition, retrieval, and generation, there are several more fine-grained single-modal sketch analysis tasks: perceptual grouping, segmentation, and parsing. These tasks need sketch analysis at the local (stroke) level. Besides their intrinsic interest, these local sketch understanding techniques can also benefit other global tasks such as sketch-based image retrieval, sketch-based video retrieval~\cite{collomosse2008free}, and sketch generation/synthesis. We next review recent advances in these areas.

\keypoint{Sketch Perceptual Grouping (SPG)} 
Humans have the ability to perceptually group visual cues  
into semantic object parts/components, which has been widely researched in Gestalt psychology area~\cite{wagemans2012century, koffka2013principles}.
Humans are able to perceptually group sketch strokes into semantic parts, \eg, airplane strokes grouped into fuselage and wings.
Thus, sketch perceptual grouping (SPG) is to imitate the human ability to group strokes into semantic parts. SPG has been studied with pre-deep learning methods ~\cite{sun2012free,qi2015im2sketch,Qi_2015_CVPR}, however progress has advanced rapidly since then. One representative application of SPG is to simplify sketches~\cite{liu2015closure}. Moreover, SPG can also be used for sketch recognition~\cite{wang2018sketchpointnet}, sketch semantic segmentation, synthesis~\cite{li2017free}, retrieval, fine-grained sketch-based image retrieval (FG-SBIR), sketch-based video retrieval~\cite{collomosse2008free}, \etc. 

Li \etal \cite{li2018universal,li2019toward} contributed the largest SPG dataset to date of $20,000$ manually-annotated sketches across $25$ object categories, and propose a universal deep grouper  
that can be applied to sketches of any category. 
Specifically, this deep universal grouper is also a sequence-to-sequence VAE with both generative and discriminative objectives:
(i) Its generative loss provides the ability to handle unseen object categories and datasets.
(ii) Its discriminative loss consists of local and global grouping losses, to guarantee both local and global consistency in the grouping outputs.

\keypoint{{Discussion}} {Shallow grouping methods mainly relied on  thresholding low-level geometric properties among the strokes, often resulting in strokes with similar geometry but different semantics being grouped. Contemporary SPG methods consider more high-level semantic and temporal information due to their deep and recurrent representations. }

\keypoint{Sketch Semantic Segmentation (SSS)} Sketch semantic segmentation has drawn attention~\cite{sun2012free,huang2014data,schneider2016example} in the free-hand sketch community as a classic topic prior to deep learning. 


Sketch semantic segmentation can potentially be addressed by conventional photo segmentation CNNs. However, these do not exploit the vector representation of strokes or their temporal patterns, loading to the development of sketch-specific segmentation models. 

Existing deep models for sketch semantic segmentation \cite{kim2018semantic,kaiyrbekov2019stroke,wu2018sketchsegnet,qi2019sketchsegnet+,wang2019spfusionnet,yang2020sketchgcn} can be grouped {according to architectures: CNN, RNN, GCN based models, \etc.} 

Li \etal \cite{li2018fast} trained a CNN-based network to transfer  well-annotated segmentation and labels from a 3D dataset to sketch domain. They used annotated 3D data~\cite{huang2014data,yi2016scalable} to produce edge-maps with partial annotations as the synthetic sketch data to train the segmentation network.

Qi \etal proposed SketchSegNet~\cite{wu2018sketchsegnet} and SketchSegNet+~\cite{qi2019sketchsegnet+}.
SketchSegNet+~\cite{qi2019sketchsegnet+} considers sketch stroke orderings and is able to process multiple object categories.
In particular, SketchSegNet and SketchSegNet+ work in an RNN-based VAE pipeline, where the Gaussian mixture model (GMM) layers of SketchRNN are replaced with  fully-connected softmax layers to predict the part labels.
Stroke-RNN~\cite{kaiyrbekov2019stroke} uses the same encoder as SketchRNN but extends the decoder to predict segmentation.

Besides the RNN backbone, other backbones have also been explored on sketch segmentation. SPFusionNet~\cite{wang2019spfusionnet} uses late fusion of CNN-RNN branches to represent sketches for segmentation. SketchGCN~\cite{yang2020sketchgcn} is a graph convolutional neural network for sketch semantic segmentation. It uses a mixed pooling block to fuse the intra-stroke and inter-stroke features from its two-branch architecture.


\keypoint{{Discussion}} {SPG essentially performs stroke-level clustering, while SSS provides stroke-level classification. \ie, SSS provides explicit part labels (category names) for each stroke, while grouping only provides aggregation relationships. SSS and SPG are analogous to the classic (supervised) semantic segmentation \cite{chen2017deeplab} and unsupervised segmentation~\cite{wang2017unsupervised} respectively \cite{li2018universal}.  Therefore, SSS needs stronger supervision during training, namely  stroke categories, in contrast to SPG's stroke grouping.}

\keypoint{{Vectorization}} {Sketch vectorization is a widely studied topic for well-drawn pencil sketches (particularly pencil-and-paper scanned sketches)~\cite{donati2017accurate,kim2018semantic,chen2018improved,guo2019deep,donati2019complete,egiazarian2020deep,stanko2020integer,parakkat2021color}; with a few studies begining to address it for free-hand sketches \cite{bhunia2021vectorization,das2020beziersketch,das2021cloud2curve}. It aims to generate vector representations for raster sketch photographs. Sketch vectorization is essentially different from sketch semantic segmentation, in that sketch vectorization aims for instance segmentation on the stroke level, rather than semantic classification for a stroke or a stroke group.}

\begin{figure}[!t]
\begin{center}
\includegraphics[width=\columnwidth]{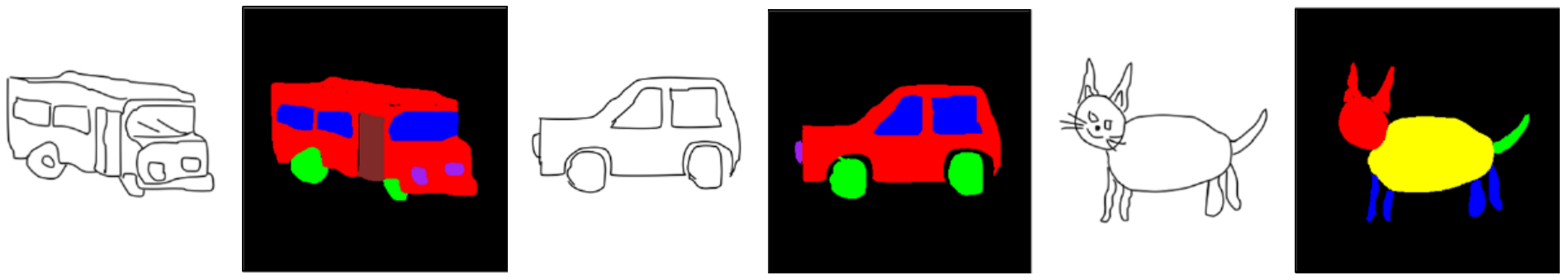}
\end{center}
   \caption{Sketches~(bus, car, cat) and ground truth annotations selected from sketch parsing paper~\cite{zheng2019deep}. The semantic parts and background are annotated by colors. Best viewed in color.}
\label{fig:parsing_samples}
\end{figure}

\keypoint{Sketch Parsing}  Recently, a new concept of  ``sketch parsing''~\cite{sarvadevabhatla2017sketchparse,jiang2019sfsegnet,mukherjeecommunicating,zheng2019deep} has gained traction.
As a kind of fine-grained semantic understanding of sketch,
sketch parsing has already been applied to assist other sketch tasks~\cite{sarvadevabhatla2017sketchparse}, \eg, sketch-based image retrieval (SBIR).
Sketch parse is related to sketch semantic segmentation. However, as shown  in Figure \ref{fig:parsing_samples}, the goal is to perform pixel-wise segmentation of the semantic regions defined by the sketch, rather than the sketch strokes as in SSS. Existing models for sketch parsing thus far only use  CNN-base networks to represent sketch the, \eg, SFSegNet~\cite{jiang2019sfsegnet} uses Deep Fully Convolutional Networks~(FCN)~\cite{long2015fully}.


\begin{figure}[!t]
\begin{center}
\includegraphics[width=0.5\textwidth]{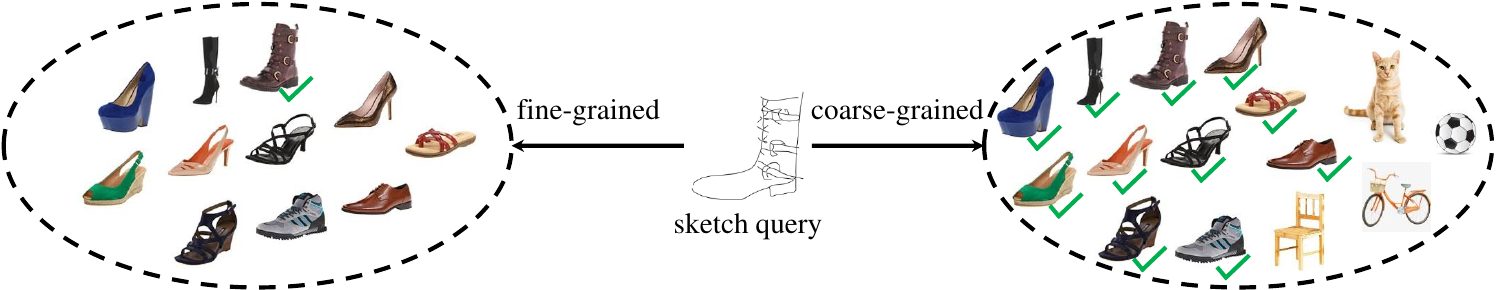}
\end{center}
   \caption{{Comparison between fine-grained (instance-level) and coarse-grained (category-level) sketch-based image retrieval. True match photos are ticked.}}
\label{fig:coarse-fine}
\end{figure}

\subsubsection{Simplification and Abstraction}
Sketch simplification has been widely studied~\cite{hilaire2006robust,chien2014line,liu2015closure,ogawa2016sketch} by the computer graphics community to simplify sketches by merging redundant strokes~\cite{ogawa2016sketch}. 
A typical pipeline~\cite{barla2005geometric} is two-stage: geometrically clustering strokes into groups (\eg, by Gestalt principles); and generating a new line to replace each group. 

{With the prevalence of deep learning, CNN-based sketch representations have been used for  sketch simplification. Simo-Serra \etal~\cite{simo2016learning} proposed a fully-convolutional network (FCN) for simplifying sketches directly from raster images of rough sketches, where a pixel level mean square error (MSE) loss is used to compare the training pairs of rough (input) and simplified (target) sketches. This approach is fully automatic and requires no user intervention. Thanks to these advantages, this FCN based model was further studied. However, this fully supervised approach needs large amounts of supervised pairs of rough sketches and their corresponding sketch simplifications as annotation. To alleviate this limitation,  Simo-Serra \etal~\cite{simo2018mastering} integrated their FCN model into the generative adversarial pipeline, where a fully-convolutional network works as the generator. This upgraded model can be jointly trained from both supervised and unsupervised data, and obtained significant performance improvements. To further study how  discriminator networks can improve sketch simplification, Xu~\etal~\cite{xu2019perceptual} proposed a multi-layer discriminator by fusing all VGG~\cite{simonyan2014very} feature layers to differentiate sketches and simplify lines, where weights used in layer fusing are automatically optimized via an intelligent adjustment mechanism. The experimental results demonstrate that this multi-layer discriminator helps the FCN based generator further improve its simplification performance. Comparing the experimental results of these three representative methods~\cite{simo2016learning,simo2018mastering,xu2019perceptual}, suggests that pixel-level losses (\ie, MSE loss) and \textit{Vanilla} discriminator loss may fail to provide adequate supervision to help  models retain semantically meaningful details when simplifying relatively complicated sketches.} 

A different take on simplification focuses on the epitome of a sketch \cite{sarvadevabhatla2015eye}. This was recently studied under the guise of 
``stroke-level sketch abstraction'' in the free-hand sketch community. Stroke-level abstraction \cite{muhammad2018learning,Muhammad_2019_ICCV} aims to abstract sketches by removing strokes that do not affect the recognizability of the sketch. Solving this problem provides several benefits: (1) It learns stroke saliency as a byproduct  \cite{muhammad2018learning} -- strokes that contribute the most to recognizability are the most salient. (2) It can be used to synthesize sketches of variable abstraction for generation, or data augmentation of discriminative sketch models \cite{muhammad2018learning}. (3) It can be used for summarization and compression more broadly \cite{Muhammad_2019_ICCV}.

The stroke-level abstraction task can be seen as a discrete combinatorial  optimization problem, and thus is intractable to solve with traditional methods. This was tackled in \cite{muhammad2018learning} by training a reinforcement learning (RL) policy to include/exclude each stroke in the sequence, while trading off between the number of included strokes and recognizability. The RL-based abstraction idea was extended by \cite{Muhammad_2019_ICCV} to re-order input strokes, rather than being constrained to the original input sequence; and further to enable customizing of the abstraction goal to preserve different aspects of `recognizability' such as category vs. attributes. 

\keypoint{{Discussion}} {Despite this good progress, simplification through \emph{merging} multiple strokes into a coarser replacement, rather than simply filtering them, remains an open question for deep learning-based sketch analysis. }

\keypoint{{Discussion}} {A bottleneck for sketch simplification is that in the existing literature the experimental results are mainly evaluated by visual comparison and user studies. Defining a good quantifiable and automatic metric to evaluate simplified sketches is an open problem and a big challenge. A good metric would be of great help in designing more well-defined loss functions for this task.}

\subsubsection{{Data Augmentations}}
\label{sec:data_augmentations}
The sketch-specific data augmentation methods discussed in this subsection can be applied to both sketch recognition and all the other sketch involved tasks (both single-modal and multi-modal), \eg, sketch-based image retrieval, sketch-related generation. 

(i) When represented as raster pictures, most common data augmentations designed for natural photos~\cite{ILSVRC15} can be applied to sketch, \eg, horizontal reflection/mirroring, rotation, horizontal shift, vertical shift, central zoom.
These augmentations have already been evaluated by the early sketch-oriented deep learning works~\cite{yusketch2015BMVC, sarvadevabhatla2015freehand}.
However, random cropping is likely unsuitable for sketch since partial sketches are often too sparse  to recognize even for humans{, and some image enhancement methods based on statistics like contrast/histogram/brightness enhancement cannot be applied to sketches.}

(ii) Stroke thickening/dilation can be used for free-hand sketch.
As discussed in some previous works on spatially-sparse convolutional neural networks~\cite{graham2014spatially}, the subtle details of sparse sketch strokes can be lost after multiple layers of convolution. Thus, this can be useful for deep neural networks that process sketches as image inputs.

(iii) Yu \etal \cite{yu2017sketch} proposed to remove the strokes to obtain more diverse sketches. Based on the observation~\cite{eitz2012humans}  that humans tend to draw outlines first before the detail, heuristics can be proposed to remove strokes with probability dependent on their sequential order.

(iv) Zheng \etal \cite{zheng2019sketch} proposed a Bezier pivot based deformation (BPD) strategy and a mean stroke reconstruction~(MSR) approach. These do not need any temporal information in the sketch. The main idea of MSR is to generate novel sketches with smaller intra-class variance. 

(v) Liu \etal \cite{liu2019unpaired} proposed two sketch-specific data augmentation strategies: (a) Manually extract some strokes from sketch SVG files to construct noise stroke masks. Then, randomly apply the noise stroke masks to the original sketches to synthesize augmented sketches.
(b) Randomly extract a patch from one sketch, and attach it to a given sketch.

(vi) Muhammad \etal \cite{muhammad2018learning} applied reinforcement learning to learn a sketch abstraction model that preserves the semantics of the sketch. Once trained, this model can be applied to generate augmentations of an input sketch at different abstraction levels.

\keypoint{{Discussion}} {Compared with augmentations on full images~(\eg, rotation, shift), the augmentation strategies above make better use of stroke information both locally and globally. However, only \cite{muhammad2018learning} makes (limited) use of human sketch variability to perform augmentation.  In future an interesting direction is to learn from the variation in sketch style between different humans, and treat sketch augmentation as a cross-human style transfer problem.}

\subsection{Multi-Modal Tasks: Sketch with Other Modalities}

Free-hand sketch has several cross-modal applications when paired with other data modalities. In this section we review sketch-related cross-modal topics including visual  (\eg, natural photo, 3D shape, video) and text domains.

Nowadays, most visual retrieval approaches work under the ``query-by-example''~(QBE)~\cite{tan1999framework} setting where users provide examples of the content that they seek. Compared with other query modalities (\eg, photo, video, text), sketch has several unique advantages:
In some scenarios users do not know the name of the object that they seek, or find it hard to describe (such as fine-grained details of a fashion item) in order to query-by-text. Meanwhile, it may be difficult or impractical to provide photos or video examples of the object that they seek. Sketch-based image retrieval provides a query modality where users express their target object by rendering their mental image in sketch. It is particularly useful when searching at the  fine-grained instance-level. Thus, sketch can be used as a modality to  retrieve natural photo, manga~\cite{matsui2015challenge}, 3D shape, video, \etc.

\subsubsection{Sketch-Photo Retrieval}
Sketch-photo retrieval is also known as sketch-based image retrieval~(SBIR)~\cite{hu2013performance,liyi2014fine,Yu_2016_CVPR,xu2016cross,xu2018crossNeurocomputing,chaudhuri2020crossatnet,yang2020instance,Fuentes_2021_CVPR,Torres_2021_CVPR}\footnote{Note that the common setting for SBIR is sketch as a query modality for images, but most methods enable either modality to be used as a query if desired. }.
{SBIR is challenging for all the reasons that sketch-analysis in general is challenging (sparse and abstract input). It is particularly challenging because of the difficulty of comparing sparse line drawings with dense pixel representations, especially when the input could be a very abstract, or iconic (symbolic) representation that is hard to compare directly to accurate perspective projection photos.}


Figure~\ref{fig:taxonomy_tree_map} includes a taxonomy for SBIR. From the perspective of evaluation criterion, SBIR can be divided into conventional/coarse-grained SBIR (\ie, category-level SBIR),  mid-grained~\cite{bui2018deep}, and fine-grained SBIR (\ie, instance-level SBIR). FG-SBIR is essentially a kind of instance-level retrieval~\cite{zheng2017sift}. From the perspective of retrieval embedding space, SBIR can be divided into common nearest-neighbor and fast hashing-based retrieval. From the perspective of supervision involved in training, SBIR can be divided into fully-supervised and zero-shot retrieval.  

\keypoint{Category and Instance Level SBIR}
In coarse-grained SBIR, given a sketch as query, a ranked list of images is returned based on the similarity (\eg, Euclidean or Hamming distance). The retrieval is judged as correct, if the photo ranked at the top has the identical class label as the query.
However, in fine-grained SBIR, the retrieval is judged as correct only when the returned photo is from the same \emph{instance} pair as the query sketch. 
{Figure~\ref{fig:coarse-fine} provides an illustration.}
Based on SBIR ideas, several sketch-based commodity search engines have been implemented, \eg, sketch-based skirt image retrieval~\cite{kondo2014sketch}, fine-grained sketch-based shoe~\cite{Yu_2016_CVPR,Song_2017_ICCV}, chair~\cite{Yu_2016_CVPR,Song_2017_ICCV}, and handbag~\cite{Song_2017_ICCV} retrieval systems. 


Some previous SBIR works~\cite{bhattacharjee2016query,lei2017sketch,bhattacharjee2018query} have used  edge-maps (image contours) of photos as an approximation to corresponding sketch images in order to perform matching. Canny edge detector~\cite{canny1986computational}, Edge Boxes toolbox~\cite{zitnick2014edge}, and holistically-nested edge detection (HED)~\cite{xie2015holistically} were usually used to extract the edges from natural photos. However, this kind of hand-designed process is now commonly replaced by end-to-end deep feature learning.

Deep sketch-based image retrieval (SBIR) has been widely studied~\cite{chopra2005learning,Yu_2016_CVPR,sangkloy2016sketchy,xu2016instance,qi2016sketch,songjifei2016deep,yu2017devil,yan2017joint,Collomosse_2017_ICCV,songjifei2017fine,huang2017deep,dey2018learning,wang2019deep,pang2019generalising,Dutta2020sSBIR} in recent years.
Existing SBIR solutions generally aim to train a joint embedding space where sketch and photo can be compared using nearest neighbor techniques. Common embedding learning approaches include:
(a) contrastive comparison based methods (implemented by pair-wise loss~\cite{wang2015spatial}), 
(b) ranking based methods~\cite{Yu_2016_CVPR,sangkloy2016sketchy},
(c) reinforcement learning based methods~\cite{bhunia2020sketch}, 
(d) deep canonical correlation analysis (DCCA)~\cite{andrew2013deep} based methods~\cite{huang2017towards}, (e) cross-domain dictionary learning~\cite{xu2018cross}, \etc.
The most widely-studied methods are ranking-based, including triplet ranking~\cite{yu2017devil,li2019sketch,Collomosse_2019_CVPR} and quadruplet ranking~\cite{seddati2017quadruplet}.


\keypoint{Ranking-Based SBIR}
We next introduce the popular triplet- and quadruplet-ranking SBIR methods in detail. 
\begin{figure}[!t]
	\centering
	\subfigure[triplet optimizing (single-modal)]{
		\includegraphics[width=0.9\columnwidth]{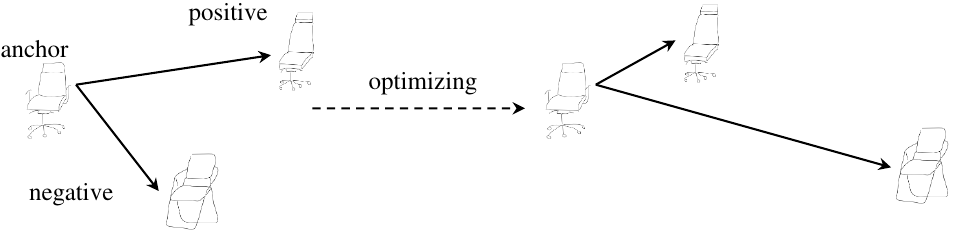}}
		
	\subfigure[triplet optimizing (multi-modal)]{
		\includegraphics[width=0.9\columnwidth]{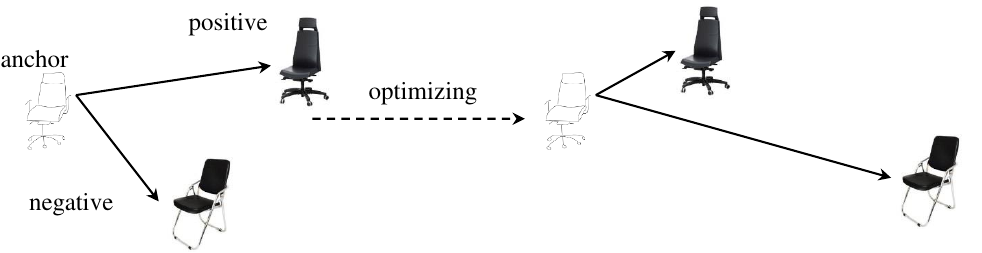}}

        \subfigure[quadruplet optimizing (single-modal)]{
		\includegraphics[width=0.9\columnwidth]{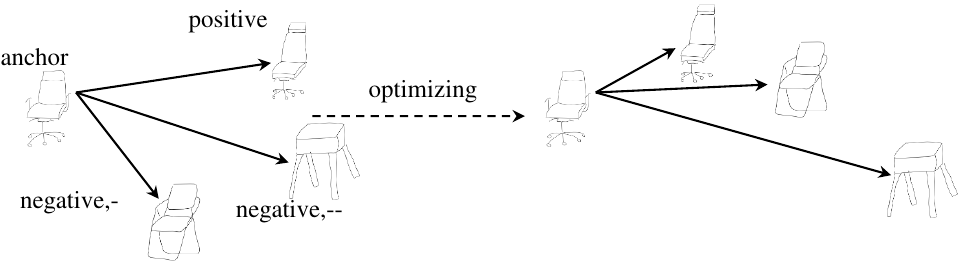}}
		
	\subfigure[quadruplet optimizing (multi-modal)]{
		\includegraphics[width=0.9\columnwidth]{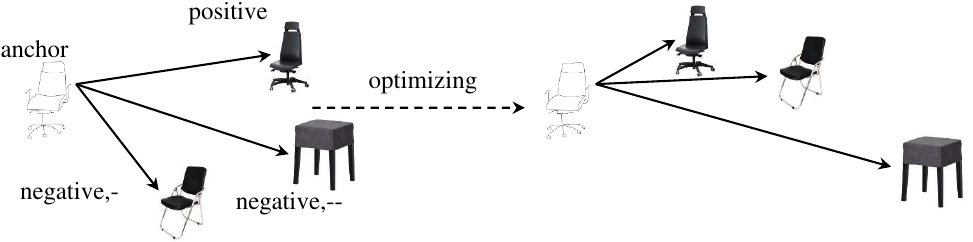}}

	\caption{ Illustration of triplet and quadruplet ranking based optimization objectives. The lengths of solid arrows illustrate the distances in embedding spaces. In the quadruplet illustration, the ``$negative,-$'' sample denotes the negative sample from the anchor category, while the ``$negative,--$'' one denotes the negative sample from the remaining categories.}
	\label{fig:triplet_quadruplet_optimizing}
\end{figure}
As shown in Figure \ref{fig:triplet_quadruplet_optimizing}, given a sketch anchor ${\bf X}_{n}$ and its positive and negative photo retrieval candidates (${\bf X}_{n,+}$, ${\bf X}_{n,-}$), the goal of triplet ranking is
\begin{equation}
\label{equ:triplet_ranking_goal}
\begin{split}
\mathcal{D}(\mathcal{F}({\bf X}_{n}), \mathcal{F}({\bf X}_{n,+})) < \mathcal{D}(\mathcal{F}({\bf X}_{n}), \mathcal{F}({\bf X}_{n,-})).
\end{split}
\end{equation}
where $\mathcal{D}(\cdot, \cdot)$ is a distance metric (\eg, $\ell_{2}$ distance).
In common FG-SBIR practice~\cite{Yu_2016_CVPR,yu2017devil}, the negative sample is usually selected from the same class as the anchor.

For quadruplet ranking~\cite{seddati2017quadruplet}, the input atom is a quadruplet of anchor ${\bf X}_{n}$, positive candidate ${\bf X}_{n,+}$, negative candidate ${\bf X}_{n,-}$ from the class of anchor, negative candidate ${\bf X}_{n,- -}$ from a different class to anchor. 
As illustrated in Figure~\ref{fig:triplet_quadruplet_optimizing}, the goal of quadruplet ranking is to ensure
\begin{equation}
\label{equ:quadruplet_ranking_goal}
\begin{split}
\mathcal{D}(\mathcal{F}({\bf X}_{n}), \mathcal{F}({\bf X}_{n,+})) < &\mathcal{D}(\mathcal{F}({\bf X}_{n}), \mathcal{F}({\bf X}_{n,-})) \\  &< \mathcal{D}(\mathcal{F}({\bf X}_{n}), \mathcal{F}({\bf X}_{n,- -})).
\end{split}
\end{equation}
Based on this, quadruplet ranking is essentially multi-task 
or multiple triplet ranking by constructing two extra triplet relationships, in order to encode more semantic information into the embedding space.
For example, Seddati \etal \cite{seddati2017quadruplet} constructed three triplets from each quadruplet, including $triplet_{a} =\{{\bf X}_{n},{\bf X}_{n,+},{\bf X}_{n,-}\}$, $triplet_{b}=\{{\bf X}_{n},{\bf X}_{n,+},{\bf X}_{n,- -}\}$ , and $triplet_{c}=\{{\bf X}_{n},{\bf X}_{n,-},{\bf X}_{n,- -}\}$.
Therefore, the quadruplet ranking loss is defined as
\begin{equation}
\label{equ:quadruplet_ranking_goal_2}
\begin{split}
\mathcal{L}_{quadruplet} = \mathcal{L}_{triplet_a} + \lambda_{b}\mathcal{L}_{triplet_b} + \lambda_{c}\mathcal{L}_{triplet_c},
\end{split}
\end{equation}
where $\lambda_{b}$, and $\lambda_{c}$ are the weights. 

In ranking-based SBIR, the anchor is usually from the sketch domain, and other samples are photos. Both triplet-ranking and quadruplet-ranking can be used for either category-level or instance-level SBIR tasks. 

\begin{figure}[!t]
\begin{center}
\includegraphics[width=\columnwidth]{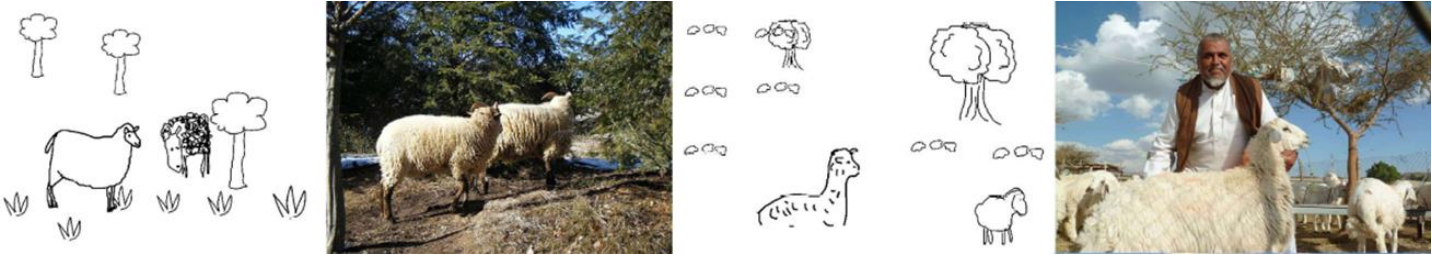}
\end{center}
   \caption{ Examples of SceneSketcher dataset~\cite{liu2020scenesketcher}, a fine-grained scene-level sketch dataset.}
\label{fig:scenesketcher-examples}
\end{figure}

{
Recently, Liu~\etal released the first scene-level fine-grained SBIR dataset, \ie, SceneSketcher~\cite{liu2020scenesketcher} (see Figure~\ref{fig:scenesketcher-examples}).
This seminal work opens a novel direction for the future SBIR research,
and contributes an effective solution to fine-grained scene sketch cross-modal matching that uses a GCN to encode the layout information of scene sketches in  fine-grained triplet ranking.
}

\keypoint{{Comment}} {
The essential principle of triplet loss is using local partial orderings to establish a global ordered relationship in the embedding space. Thus triplet ranking~\cite{schroff2015facenet} can be understood as Topological Sorting\footnote{\url{https://en.wikipedia.org/wiki/Topological_sorting}}.
The triplet annotations work as a partially ordered set.
Compared with other loss functions, the main advantages of triplet loss are:
(i) It helps to involve more local partial orderings and annotations to learn more fine-grained embedding space.
(ii) Given a limited number of $N$ training samples, their triplet orderings have $C_{N}^{3}$ combinations, producing significant annotation augmentation. This is beneficial for training deeper networks on smaller sketch datasets.
It should be noted that the performance of triplet loss is heavily dependent on (a) the choice of margin parameter and (b) the triplet construction strategy.  
}

\keypoint{{Comment}} {
We remark that ranking-based  SBIR models can also be improved by multi-task training along with classification~\cite{sangkloy2016sketchy, bui2016generalisation,lin2019tc}. Furthermore, rather than purely discriminative training, SBIR can also be tackled by generating one modality from the other \cite{guo2017sketch}, \eg, using conditional GAN \cite{radford2015unsupervised}; or using generative losses to regularize discriminative training \cite{pang2017cross}. SBIR training can also be combined with post-processing re-ranking~\cite{bhattacharjee2016query,bhattacharjee2018query,huang2018sketch} to refine the initially learned embedding spaces.}

\begin{figure}[!t]
\begin{center}
\includegraphics[width=\columnwidth]{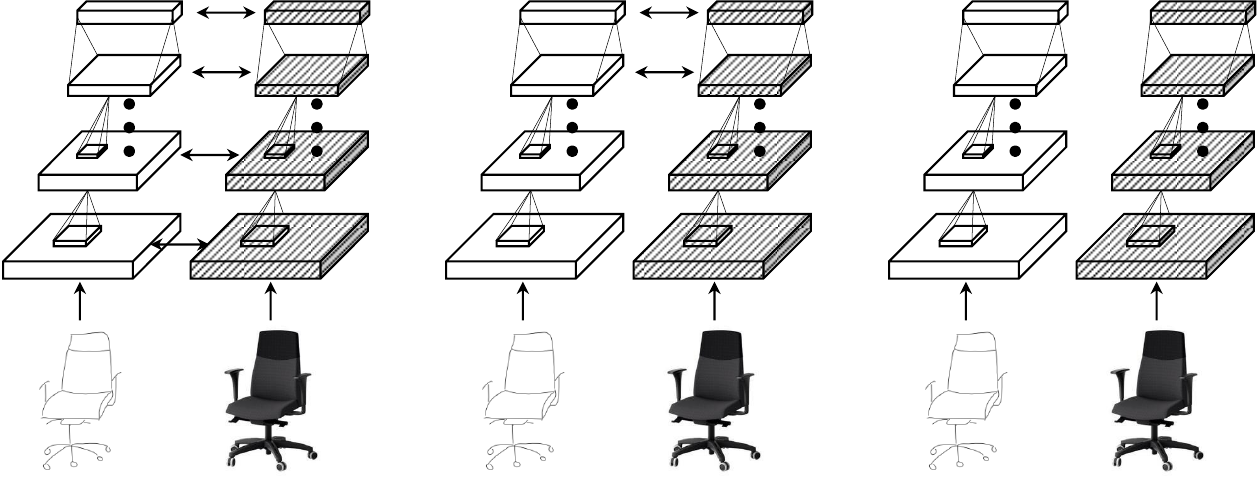}
\end{center}
   \caption{Different weight sharing manners (left: Siamese, middle: semi-heterogeneous, right: heterogeneous) for CNN-based cross-modal networks. The hollow and shaded networks denote the branches for sketches and photos, respectively. The double sided arrows indicate sharing of weights.}
\label{fig:triplet_architectures}
\end{figure}

\keypoint{Network Architectures in SBIR} SBIR models generally need two or more branches to process sketches and photos for comparison using the metrics introduced above. As shown in Figure~\ref{fig:triplet_architectures}, both triplet and quadruplet ranking models can use backbone networks that are  Siamese,  semi-heterogeneous, or heterogeneous. 
(i) Siamese networks~\cite{Yu_2016_CVPR} use full weight-parameter sharing across branches.
(ii) Semi-heterogeneous network~\cite{bui2017compact,lei2019semi} use partial weight sharing across the branches. Typically early layers are modality-specific, and weight-tied layers are at deeper layers. 
(iii) In heterogeneous networks~\cite{zhang2017sketch}, the sketch branch uses independent parameters to the photo branches. 
The trade-offs underlying these architectures are that sharing weights enables more data (both sketch and photo) to be used to estimate parameters, reducing overfitting. But separating weights enables the sketch/photo branches to adapt more specifically to their respective domains. Weight sharing considerations are discussed in more detail in~\cite{bui2016generalisation}.


\keypoint{Hashing-Based SBIR}
In order to achieve faster sketch-based image retrieval, recent research has studied optimizing the feature coding (\eg, sketch-image hashing~\cite{Liu_2017_CVPR,Zhang_2018_ECCV}), and the feature map (\eg, Asymmetric Feature Maps~\cite{Tolias_2017_CVPR}).

In particular, sketch-image hashing (or hashing SBIR) has gained attention.
Liu \etal \cite{Liu_2017_CVPR}, propose the first deep hashing model for SBIR, which is  a classic deep hashing pipeline including: (i) feature extractor network, (ii) hashing layer with binary constraints, and (iii) hashing loss. 
This classic pipeline 
has been widely studied in  photo-oriented deep hashing~\cite{lin2015deep,wang2017survey}, where the hashing layer is typically fully-connected with sigmoid or tanh activation, and a discrete binary constraint. The loss functions of deep hashing models are often non-differentiable, due to the discrete binary constraints. Thus, common practice is that the feature extractor backbone and hashing layer are alternatively optimized in two separate steps by fixing one and optimizing the other.


Existing SBIR hashing models work on the SBIR benchmarks, \ie, Sketchy~\cite{sangkloy2016sketchy} (75K sketches) and TU-Berlin Extended~\cite{Zhang_2016_CVPR} (20K). The scale of these benchmarks is not yet large enough to thoroughly test  hashing SBIR methods. 

\keypoint{{Discussion}} {Current issues in SBIR include: Self-supervised pre-training for SBIR \cite{pang2020jigsaw,bhunia2021vectorization}, optimizing SBIR for early retrieval using partially drawn sketches, for example using reinforcement learning \cite{bhunia2020sketch}; investigating whether costly sketch-photo annotation pairs can be replaced with edge-maps \cite{radenovic2018dsm} and cross-category generalization of SBIR which is discussed next.}

\subsubsection{Zero-Shot SBIR}
Many existing SBIR works assume that categories to be queried are included in the training set. In recent years, motivated by the zero-shot setting for supervised photo retrieval~\cite{sablayrolles2017should}, zero-shot sketch-based image retrieval (ZS-SBIR) has also been studied~\cite{lu2018learning,thong2019open,duttastyle2019,dey2019doodle,dutta2019semantically,Yelamarthi_2018_ECCV,li2019bi,pandey2019adversarial,kumar2019generative,Liu_2019_ICCV,pandey2020stacked,xu2020progressive,pang2019generalising,chaudhuri2020simplified,zhang2020zero,chaudhuri2020zero}. Similar to natural photo zero-shot learning/recognition~\cite{fu2015transductive,fu2018recent,wang2019survey},  ZS-SBIR systems aim to enable query and retrieval of categories that are from  \textit{unseen} categories. \ie, categories that have not been involved in training stage. This is important in practice, \eg, for an e-commerce application of SBIR, where new products should ideally be enrolled in the search engine without requiring re-training. 

\keypoint{{Discussion}} {
{ZS-SBIR systems can follow conventional zero-shot learning methods \cite{fu2015transductive,fu2018recent,wang2019survey} in exploiting auxiliary knowledge such as word vectors~\cite{mikolov2013distributed},  attributes~\cite{lampert2013attribute}, or class hierarchy to define the model for the unseen class. However, directly synthesizing a retrieval model for novel-classes with auxiliary knowledge leads to the same challenges of ZSL (cross-category domain-shift \cite{fu2015transductive} and inconvenient need to specify nameable categories at testing-time \cite{fu2015transductive}). Meanwhile, it would entail new challenges specific to SBIR: (i) Knowledge transfer needs to occur across both sketch and photo views. (ii) Some kinds of auxiliary knowledge may not make sense for sketch (\eg, \emph{banana-is-yellow} may be visible in photo but not sketch). Meanwhile, auxiliary knowledge transfer is not strictly necessary for retrieval in the way that it is for category recognition.  Therefore many ZS-SBIR methods tackle the problem in a 
\emph{domain generalization} \cite{Li_2017_ICCV} manner. That is, training a matching network on the training categories that is robust enough to support direct application to unseen testing categories. }
}
{
Thus, common approaches are to train ranking \cite{dey2019doodle,pandey2020stacked}  or generative \cite{dutta2019semantically,Yelamarthi_2018_ECCV} models for retrieval, which are enhanced and made robust by constraints such as domain-alignment losses \cite{dey2019doodle,pandey2020stacked,dutta2019semantically} and auxiliary semantic knowledge reconstruction \cite{dey2019doodle,dutta2019semantically}. In these cases the auxiliary semantic knowledge is only used to constrain representation learning at train time and is not required during testing time as for conventional ZSL -- thus maintaining the vision that SBIR should only depend on ability to depict and not to verbally describe. 
}

{
Current directions include extending SBIR to the generalized zero-shot setting, where testing categories are a mix of training and unseen categories \cite{pandey2020stacked,dutta2019semantically};  extending sketch-photo hashing to the zero-shot setting~\cite{Shen_2018_CVPR}; and training SBIR without paired samples \cite{duttastyle2019}.}

\subsubsection{{Sketch-Photo Generation}}
\label{sec:sketch_photo_generation}

\begin{table*}[!t]
\caption{{Comparison of the representative pipelines of deep sketch-photo generation. ``s $\rightarrow$ p'' and ``p $\rightarrow$ s'' denote ``sketch $\rightarrow$ photo'' and ``photo $\rightarrow$ sketch'', respectively.} {The backbones of the representative references here are implemented by CNNs.}}
\label{table:sketch-photo-generation}
\begin{center}
\resizebox{\textwidth}{!}{
\begin{tabular}{ c | c | c | c | l}
\hline
Tasks & {Pipelines} &  Representative Ref.  & Sketch-Specific Designs & (Dis)Advantages \\
\hline
\hline


%
%
%
%
%

s $\rightarrow$ p &       GAN            &      \cite{Chen_2018_CVPR,Jo_2019_ICCV,yang2020deep}                 & Sketch-specific designs are generally injected in the generators.  & \tabincell{l}{A: simple, end-to-end training  \\ D:     suboptimal performance        }  \\   

s $\rightarrow$ p &      GAN \#1 $\rightarrow$ GAN \#2            &       \cite{xia2019cali,Ghosh_2019_ICCV}                & GAN \#1 for stroke refinement, GAN \#2 for  photo synthesis        &    \tabincell{l}{A: clear motivation  \\ D:   multi-stage training          }   \\ 

s $\rightarrow$ p &        CGAN           &        \cite{li2020staged,liu2020sketch}   &  Specific designs and domain knowledge can be injected as  conditions.         &   \tabincell{l}{A: clear motivation    \\ D:    sensitive to conditions         }  \\ 

s $\rightarrow$ p &       ContextualGAN            &      \cite{Lu_2018_ECCV}                 &  learns joint distribution of sketch-photo pairs         &    \tabincell{l}{A: appearance freedom, less strict alignment    \\ D:   multi-stage training          }  \\ 

s $\rightarrow$ p &  TextureGAN    &          \cite{xian2018texturegan}             &   supports  local texture constraints     &  \tabincell{l}{A:  more fine-grained   \\ D:      high complexity       }   \\   

\hline
p $\rightarrow$ s &   CGAN &  \cite{li2019photo}    &   Specific designs and domain knowledge can be injected as  conditions.   &  \tabincell{l}{A: clear motivation    \\ D:    sensitive to conditions         } \\
p $\rightarrow$ s &   conditional encoder-decoder &  \cite{kampelmuhler2020synthesizing}    &    Conditions a convolutional decoder on a class prior.  &  \tabincell{l}{A:  simple, end-to-end training  \\ D:    suboptimal performance           }  \\
{p $\rightarrow$ s} & {VAE (CNN encoder + RNN decoder) }   &  \cite{song2018learning}    & {shortcut cycle consistency, RNN decoder} & {\tabincell{l}{A:  stroke by stroke, end-to-end training  \\ D:    suboptimal for long strokes           }}   \\

\hline
\end{tabular}
}
\end{center}

\end{table*}

Sketch and photo based mutual generation (translation/synthesis) is a classic cross-modal topic of sketch research covering both:
(i) sketch-to-photo generation~\cite{Gao2020SketchyCOCO,chen2020deepfacedrawing,huang2020multi},
(ii) photo-to-sketch generation~\cite{song2018learning,li2019photo,zhang2019unpaired,kampelmuhler2020synthesizing}.
In particular, sketch-to-photo generation methods have addressed: 
(a) sketch to photo~\cite{li2020staged}, 
(b) sketch \& photo to photo~\cite{portenier2018faceshop,Jo_2019_ICCV},
(c) sketch/edge \& color to photo~\cite{Sangkloy_2017_CVPR}.
Sketch and photo based generation can be used to help users to create or design novel images in various practical applications: sketch-based photo editing~\cite{portenier2018faceshop,Jo_2019_ICCV,yang2020deep}, sketch to painting generation~\cite{li2018line},
cloth design~\cite{li2018foldsketch,wang2018learning}, sketch to natural photo generation~\cite{Ghosh_2019_ICCV}, \etc.
In some cases, sketch-photo generation also involves style transfer~\cite{collomosse2017sketching,liu2020sketch}. 

Note that:
(i) Sketch-to-photo generation aims to solve  cross-modal translation from abstract and sparse line drawings to pixel space, different to well-drawn sketch colorization~\cite{zou2018lucss,zhang2018two,zou2019language}.
(ii) Photo-to-sketch generation does not refer to extracting edge-map from natural photos~\cite{zitnick2014edge,soria2019dense} (edge-maps of literal perspective projections), but instead needs models that learn to mimic human sketching and abstract drawing style \cite{song2018learning}. 

\begin{figure}[!t]
\begin{center}
\includegraphics[width=0.7\columnwidth]{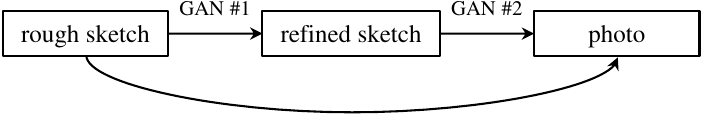}
\end{center}
   \caption{Illustration of two-stage two-GAN based sketch-to-photo cross-modal generation idea~\cite{xia2019cali,Ghosh_2019_ICCV}. The arrow lengths denote the distances of cross-domain gap. See text for details.}
\label{fig:two-gans}
\end{figure}

Sketch and photo generation methods have been widely studied~\cite{zhu2017unpaired,isola2017image,Chen_2018_CVPR,xia2019cali,liu2019unpaired} based on various GAN~\cite{goodfellow2014generative} variants including 
conditional GAN~\cite{mirza2014conditional}, cycle GAN~\cite{zhu2017unpaired}, and texture GAN~\cite{xian2018texturegan}. 
{
Meanwhile, VAE also works well for sketch-photo generation.
We compare the existing deep sketch-photo generation pipelines and their representative models in Table~\ref{table:sketch-photo-generation}.
The domain gap from sketch to photo is large.
As demonstrated in Figure~\ref{fig:two-gans}, an intuitive idea is to decompose the large gap into two smaller gaps.
Some previous works~\cite{xia2019cali,Ghosh_2019_ICCV} proposed to use two GANs to achieve sketch-to-photo generation in two steps:
(i) use the first GAN to refine rough sketches, \eg, generating good contours~\cite{Ghosh_2019_ICCV}, and
(ii) input the refined sketches into the second GAN to generate the target photos. 
}

\keypoint{{Discussion}} {
Sketch-photo cross-modal generation is distinctively different from conventional photo-to-photo translation \cite{isola2017image}. Existing photo-to-photo models can  assume pixel-wise alignment between inputs and outputs. However, this requirement is strongly violated in the case of sketches and photos. Indeed no simple warping can provide pixel-wise alignment between sketch and photo given the potentially  abstract or iconic nature of sketches. Thus, existing sketch-photo synthesis work has made efforts to work around this issue, \eg, using contextual GAN \cite{Lu_2018_ECCV}.  
}

\keypoint{{Discussion}} {
Generating sketch images as pixels suffers from the problem that blurriness in an image that should be made of sharp edges is very visible. However, an important difference between sketch-photo translation and photo-photo translation is that the raw format of sketch data is often a time-series of way-points. If such raw sketch representation is to be used, the encoder or decoder should be an RNN rather than CNN. The first example of such sequential vectorized photo-sketch translation is in \cite{song2018learning}, where sharp sketches are sequentially produced using a recurrent decoder. 
}

\keypoint{{Discussion}} {Other current considerations are similar to that in image-image translation including building sketch-photo translation models that work based on unpaired samples. }

\keypoint{{Discussion}} {Sketch inpainting is an interesting task related to sketch generation, with a wide range of practical applications in engineering~\cite{sasaki2017joint,chen2019blind}, medicine~\cite{chen2019blind}, agriculture~\cite{chen2018root,chen2019blind,chen2019adversarial}, \etc. This task can be considered as a single-modal generation task from deteriorated sketch (with discontinuities strokes) to a restored complete sketch. \eg,  old sketches~\cite{SasakiCGI2018} and engineering sketches~\cite{sasaki2017joint}. Sketch inpainting is also studied for generating/parsing the sketch structures from non-sketch images (\eg, retina~\cite{chen2019blind}, plant roots~\cite{chen2018root,chen2019blind,chen2019adversarial}, road networks~\cite{chen2019blind} from satellites), which is a cross-modal photo-to-sketch generation task.  How to solve this problem in end-to-end framework is still challenging and under-studied.}


\subsubsection{{Sketch-3D Retrieval}}
Sketch-3D retrieval refers to using sketch as query to retrieve 3D  models~\cite{shao2011discriminative}. {Compared to SBIR, 3D retrieval is more challenging due to the larger domain gap between 2D sketch and 3D model. }

\begin{figure}[!t]
\begin{center}
\includegraphics[width=\columnwidth]{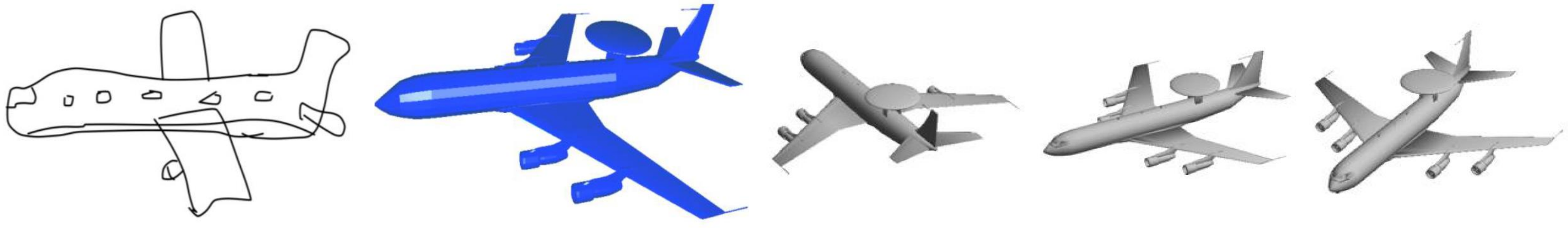}
\end{center}
   \caption{Illustration of view matching across sketch and 3D shape. Images (from left to right: sketch, 3D shape, three random views of the 3D shape) are selected from~\cite{li2017multi}.}
\label{fig:sketch_3D_retrieval}
\end{figure}

Sketch-3D retrieval was studied well before the deep learning era \cite{li2014comparison}. Classic approaches often proceeded in a two-stage manner \cite{li2016semantic}: (i) View selection: Use an automatic procedure to select representative viewpoints of a given 3D model, hoping that one of the selected viewpoints is similar to that of the query sketches; and (ii) Projection and Matching: Project each selected view of the 3D model into 2D space by line rendering algorithm~\cite{decarlo2003suggestive}. Then match the sketch against the 2D projections of the model based on pre-defined features such as SIFT~\cite{lowe2004distinctive}.
See Figure~\ref{fig:sketch_3D_retrieval} for an illustration. However, as argued in some previous work~\cite{Wang_2015_CVPR},  view selection is a bottleneck of the two-stage approach as the ``best'' views are subjective and ambiguous. Moreover, matching based on hand-crafted features is inaccurate.

Gradually, sketch based 3D model retrieval has been studied within the end-to-end deep learning paradigm~\cite{Wang_2015_CVPR,zhu2016learning,ye20163d,li2017multi,Chen_2018_ECCV,Chen_2019_CVPR,qi2021fg3dsbir}.
As a representative method, Wang \etal \cite{Wang_2015_CVPR} proposed to use two Siamese networks to learn the sketch and projected views directly in the end-to-end manner, which takes a quadruplet of sketches and projected viewpoints as input, and uses multiple pair-wise losses.
In each quadruplet, two sketches (${\bf X}_{1}$, ${\bf X}_{2}$) and two viewpoints (${\bf V}_{1}$, ${\bf V}_{2}$) are randomly selected from sketch and 3D domains, respectively.
For simplicity they assume that ${\bf X}_{1}$ and ${\bf X}_{2}$ are from the same category sharing a Siamese network, while ${\bf V}_{1}$ and ${\bf V}_{2}$ are also from the same category sharing another Siamese network.
The  quadruplet loss is
\begin{equation}
\label{equ:wang_2015_cvpr_loss}
\begin{split}
\mathcal{L}({\bf X}_{1}, {\bf X}_{2}, & {\bf V}_{1}, {\bf V}_{2}) = \mathcal{L}_{pair}({\bf X}_{1}, {\bf X}_{2}) \\ 
&+ \mathcal{L}_{pair}({\bf V}_{1}, {\bf V}_{2}) + \mathcal{L}_{pair}({\bf X}_{1}, {\bf V}_{1}),
\end{split}
\end{equation}

\noindent The loss function terms  
$\mathcal{L}_{pair}({\bf X}_{1}, {\bf X}_{2})$ and $\mathcal{L}_{pair}({\bf V}_{1}, {\bf V}_{2})$ enable the network to learn category-level similarity within each domain; while the $\mathcal{L}_{pair}({\bf X}_{1}, {\bf V}_{1})$ term forces the network to learn cross-modal similarity. 
Given input samples $a$ and $b$, the pair-wise loss function is defined as:
\begin{equation}
\label{equ:wang_2015_cvpr_pair_loss}
\begin{split}
\mathcal{L}_{pair}(a, b) = \left\{
\begin{aligned}
\alpha {\mathcal{D}(\mathcal{F}_{a}(a), \mathcal{F}_{b}(b))}, & ~~\textrm{ if } y_{a} \neq y_{b},\\
\beta e^{\gamma \mathcal{D}(\mathcal{F}_{a}(a), \mathcal{F}_{b}(b))}, & ~~\textrm{ otherwise },
\end{aligned}
\right.
\end{split}
\end{equation}
where $y_{a}$ and $y_{b}$ are the corresponding class labels, and $\mathcal{F}_{a}(\cdot)$ and $\mathcal{F}_{b}(\cdot)$ denote the feature extractions that have been applied to $a$ and $b$, respectively. 

Besides this pair-wise deep metric learning, other deep metric learning methods also can be applied to Sketch-3D matching, \eg, triplet ranking~\cite{qi2018semantic,kuwabara2019query} and deep correlation metric learning~\cite{dai2017deep,dai2018deep}. 

Moreover, some previous works also studied how to represent 3D models more comprehensively in sketch based 3D retrieval tasks.
For instance, Xie \etal \cite{Xie_2017_CVPR} proposed to represent 3D models by computing the Wasserstein distance~\cite{bogachev2012monge} based barycenters of multiple projections of 3D models.


\subsubsection{{Sketch-3D Generation}}
Sketch to 3D model generation is also an interesting cross-modal research topic analogous to the sketch-to-image generation discussed in Section~\ref{sec:sketch_photo_generation}. 
Using sketch to generate 3D models/shapes~\cite{wang2018unsupervised,Zhang_2021_CVPR} is extremely challenging but has important applications such as sketch-based product design~\cite{shen2019deepsketchhair,Willis_2021_CVPR}. Compared to the other tasks discussed, this is relatively under-studied thus far. 
Most of the existing deep learning based sketch-to-3D generation models are engineered for highly well-drawn  or professional pencil sketches~\cite{huang2016shape,han2017deepsketch2face}.
Recently, 3D-to-sketch~\cite{ye2019deepshapesketch} generation has also been explored in a deep learning manner. 

\begin{figure}[!t]
\begin{center}
\includegraphics[width=0.9\columnwidth]{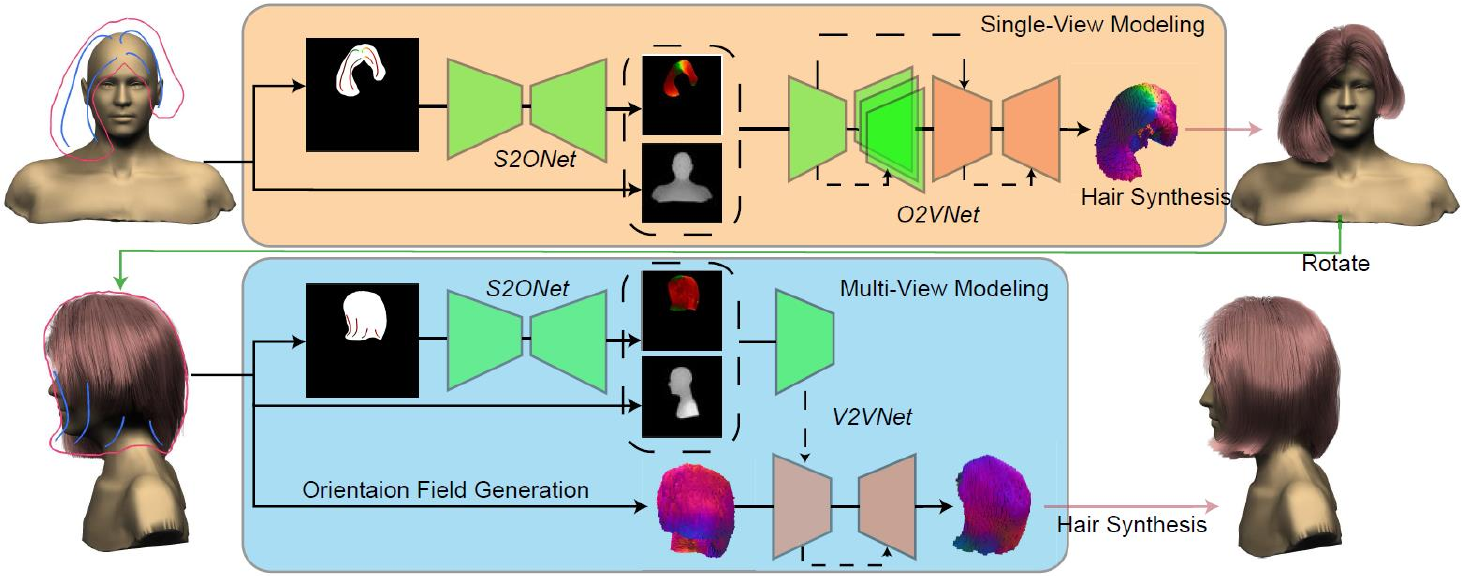}
\end{center}
   \caption{Pipeline of a sketch-based 3D hair modeling deep model~\cite{shen2019deepsketchhair}. This system takes 2D sketch as input and generates a realistic 3D hairstyle.}
\label{fig:deep-sketch-hair}
\end{figure}

{
\textit{DeepSketchHair}~\cite{shen2019deepsketchhair} is a representative deep model that generates realistic 3D hairstyle models from 2D sketches.
As shown in Figure~\ref{fig:deep-sketch-hair}, given a 3D bust model as a reference, the system takes in a user-drawn sketch (consisting of hair contour (red lines) and a few strokes indicating the hair growing directions (blue lines)  within a hair region), and automatically generates a 3D hair model, which matches the input sketch both globally (for contour) and locally (for growing directions).
This model solves two challenging cross-domain mappings:
(i) mapping sketch to dense 2D hair orientation field, by S2ONet,
and (ii) mapping 2D orientation field to 3D vector field, by O2VNet.
}

\subsubsection{{Sketch-Video Retrieval}}

Sketch based video retrieval~(SBVR) has been studied~\cite{collomosse2009storyboard} prior to contemporary deep learning. 
{SBVR is also highly challenging due to the huge domain gap between free-hand sketch and video.}
In SBVR applications using sketch as query has the advantage that humans can use lines or arrow vectors to describe moving or other dynamic scenes.
Thus, sketch can be used to not only depict static objects and scenes, but also motion information.


{Performing SBVR using sketches conveying both appearance and motion  is challenging due to the need to segment the motion and appearance information from the sketch and use it to address the corresponding channels in the video gallery for retrieval.} To address this challenge, a multi-stream multi-modality deep network was proposed in ~\cite{xu2020fine}. This study further extended SBVR to fine-grained SBVR to perform instance-level retrieval of videos given sketch queries.

Finally, we note that, motion sketch based crowd video retrieval ~\cite{wu2016motion,wu2017motion} has been studied recently, which is useful for video surveillance analysis.

\keypoint{{Discussion}} {Which kinds of videos are appropriate and feasible for sketch-based retrieval is still an open question. As are human-computer-interaction questions around how time, shot change, and motions should be depicted in sketch.}

\subsubsection{{Other Sketch-Related Multi-Modal Tasks}}

Recently some other interesting sketch based multi-modal tasks have emerged, \eg, text-to-sketch generation~\cite{huang2019sketchforme} (\eg, text instruction based conversational authoring of sketches~\cite{huang2020scones}),
sketch-based photo classifier generation~\cite{Hu_2018_CVPR}, sketch-based segmentation model generation \cite{hu2020sas}, 
sketch-based pictionary games~\cite{sarvadevabhatla2018game,sarvadevabhatla2020pictionary,bhunia2020pixelor}, sketch to photo contour transfer~\cite{pang2018deep}, 
and {sketch-guided object localization in natural photos~\cite{tripathi2020sketch}}.

%

\subsection{Experimental Comparison}\label{sec:comparison}
\subsubsection{Representing Sketch}
{As discussed in Section~\ref{sec:Intrinsic_Traits_and_Domain-Unique_Challenges}, sketch can be represented in several diverse formats such as raster image, and waypoint sequence. These representations lend themselves to different neural network architectures. We therefore take the opportunity to use \torchsketch{} to perform the first thorough comparison of neural network architectures for sketch recognition/representing. } 

\begin{table*}[!tb]
\caption{
Comparison of recognition accuracy for different network architectures on a subset~\cite{xu2019multi} of QuickDraw~\cite{ha2018sketchrnn}. }\label{table:comparison_on_recognition}
\begin{center}
\resizebox{\textwidth}{!}{
\begin{tabular}{l | l | c | c | c | c | r }
\hline
 \multicolumn{2}{c|}{\multirow{2}{*}{Architecture \& Network}}  & \multirow{2}{*}{Input} & \multicolumn{3}{c|}{Recognition Accuracy} & \multirow{2}{*}{\tabincell{c}{Parameter \\ Amount}}  \\
  \cline{4-6}
\multicolumn{2}{c|}{}   & & acc.@1 & acc.@5 & acc.@10 &   \\
  \hline
\multirow{17}{*}{Convolutional Neural Networks (CNNs)} & AlexNet~\cite{krizhevsky2012imagenet} & \multirow{17}{*}{picture} & {0.6808} & {0.8847} & {0.9203} & 58,417,305  \\

& VGG-11~\cite{simonyan2014very} &   & {0.6743} & {0.8814} & {0.9191} & 130,179,801 \\

& VGG-13~\cite{simonyan2014very} &   &  0.6808  & 0.8881 & 0.9232 &  130,364,313 \\

& VGG-16~\cite{simonyan2014very} &   & 0.6837 & 0.8889  & 0.9253 &  135,674,009 \\

& VGG-19~\cite{simonyan2014very} &   &  0.6908 & 0.8839  & 0.9208 &  140,983,705 \\

& Inception V3~\cite{szegedy2016rethinking} & &  {0.7422} & {0.9189} & {0.9437}  & 25,315,474  \\

& ResNet-18~\cite{he2016deep} &  & {0.7164} & {0.9072} & {0.9381}  & 11,353,497  \\

& ResNet-34~\cite{he2016deep} &   & {0.7154} & {0.9083} & {0.9375}  & 21,461,657  \\

& ResNet-50~\cite{he2016deep} &   & 0.7043 & 0.8987 & 0.9303  &  24,214,937  \\

& ResNet-101~\cite{he2016deep} &   & 0.7071  & 0.8992 & 0.9317 &  43,207,065  \\

& ResNet-152~\cite{he2016deep} &  & {0.6924} & {0.8973} & {0.9312} & 58,850,713 \\

& DenseNet-161~\cite{huang2017densely} & &  0.7008 & 0.8971 & 0.9302 & 27,234,105   \\

& DenseNet-169~\cite{huang2017densely} & &  0.7173 & 0.9050 & 0.9358 & 13,058,905  \\

& DenseNet-201~\cite{huang2017densely} & &  {0.7050} & {0.9013} & {0.9331} & 18,755,673  \\

& MobileNet V2~\cite{Sandler_2018_CVPR} & &   {0.7310} & {0.9161} & {0.9429} & 2,665,817 \\

\hline
\multirow{2}{*}{Recurrent Neural Networks (RNNs)} & LSTM & \multirow{4}{*}{stroke vector} &  0.6068 & 0.8416 & 0.8931 &  2,593,881  \\

& Bi-directional LSTM & & {0.6665} & {0.8820} & {0.9189} & 5,553,241  \\

& GRU & & 0.6224 & 0.8574 & 0.9055 & 2,000,473  \\

& Bi-directional GRU & & {0.6768} & {0.8854} & {0.9234} & 5,419,097  \\

 \hline
\multirow{5}{*}{Graph Neural Networks (GNNs)} &  Graph Convolutional Network~(GCN)~\cite{kipf2017semi} & \multirow{5}{*}{stroke vector}  & {0.6800}  & {0.8869}  & {0.9224}  & 6,948,441    \\
& Graph Attention Network~(GAT)~\cite{velickovic2018graph} & & {0.6977} & {0.8952}  & {0.9298}   &   11,660,889    \\
& \textit{Vanilla} Transformer \cite{vaswani2017attention} &  & {0.5249} & {0.7802} & {0.8486} & 14,029,401   \\
& Multi-Graph Transformer (Base)~\cite{xu2019multi} &  & {0.7070} & {0.9030} & {0.9351} & 10,096,601  \\
& Multi-Graph Transformer (Large)~\cite{xu2019multi} &  & {0.7280} & {0.9106} & {0.9387} & 39,984,729 \\
\hline
Textual Convolutional Network (TCN) & TCN \cite{xu2020deep} & stroke vector &  0.5511  & 0.8020  &  0.8646  &  2,750,873   \\

\hline
\end{tabular}
}
\end{center}
\end{table*}

To this end we follow~\cite{xu2019multi} in using 414K sketches drawn from QuickDraw. These are organized into training, validation, and test splits composed of 1000, 100, and 100 sketches respectively from each of the 345 QuickDraw categories.
Following~\cite{xu2018sketchmate}, we truncate or pad sketch samples to a uniform length of 100 key points/steps to facilitate efficient training of RNN-, GNN-, and TCN-based models, where each time-step is a $4D$ input (\ie, two coordinates and two pen state bits). Sketch recognition is a fundamental topic within the field, and the QuickDraw sketches are subject to realistic abstraction, noise, and drawing diversity. We therefore hope that this benchmark can help practitioners while supporting future research in the field. 

The recognition accuracy across these architectures, as implemented by \torchsketch,  are reported in Table~\ref{table:comparison_on_recognition}. We can analyze these results with comparisons within and across architecture categories. 
Within each architecture, we can observe from Table~\ref{table:comparison_on_recognition}:
(i) For CNNs, the deeper networks (\eg, DenseNet-161) have no obvious advantage compared to the shallower networks (\eg, AlexNet, VGG). This is likely due to the sparsity of sketch where redundant convolution and pooling operations lose information about the sparse pixels. 
(ii) For RNNs, bidirectional networks outperform the unidirectional networks by a clear margin (0.66+ vs. 0.60+). This makes sense as human sketch ordering is only loosely consistent \cite{li2017free}. 
(iii) For GNNs, multi-graph transformer~(MGT) outperforms graph convolutional network~(GCN) and graph attention network~(GAT).

Across the architectures, we can observe:  
(i) Thus far the best CNN networks (InceptionV3) outperform the best sequential networks. This may be because the sequential networks (RNN, GNN, and TCN) truncate the input coordinate sequences.
(ii) Among  GNNs, the multi-graph transformer \cite{xu2019multi} comes closest to matching peak CNN performance. (iii) 
Compared with CNNs and GNNs, RNNs and TCN have significantly fewer parameters. However (iv) TCN, performs unsatisfactorily in this fully-supervised setting.

%

\begin{table*}[!t]
\caption{{Robustness of ResNet-18 CNN backbone to perturbations in the form of random spatial transformations. \\
Each setting is tested $10$ times, and mean ($\%$) and standard deviation of performance are reported.}}
\label{table:robustness}
\begin{center}
\resizebox{\textwidth}{!}{
\begin{tabular}{ l | c  c  c  c  c  c  c  c  c  c}
\hline
\multirow{2}{*}{\tabincell{c}{Spatial \\ Transform}} &  \multicolumn{3}{c|}{Recognition on TU-Berlin} & \multicolumn{4}{c|}{\tabincell{c}{Coarse-Grained SBIR \\ on TU-Berlin Extended}} &  \multicolumn{3}{c}{\tabincell{c}{Fine-Grained SBIR \\ on QMUL Shoe}} \\
\cline{2-11}
& acc.@1  & acc.@5 & acc.@10 & rank@1 & rank@5 & rank@10 & mAP & rank@1 & rank@5 & rank@10 \\
\hline
shift &         67.16$\pm$0.31             &            88.20$\pm$0.24         &            92.82$\pm$0.32               & 51.17$\pm$0.36 & 71.04$\pm$0.31 & 77.33$\pm$0.21 & 28.91$\pm$0.05 & 21.13$\pm$1.42 & 50.17$\pm$2.42 & 67.39$\pm$2.10 \\
scale &            44.07$\pm$0.80             &             68.49$\pm$0.62                 &             77.08$\pm$0.48              & 31.24$\pm$0.33 &  51.02$\pm$0.56 &59.06$\pm$0.50 &15.75$\pm$0.24 &8.78$\pm$2.15 &25.74$\pm$2.10 &39.65$\pm$3.02 \\
horizontal flip &                 70.60$\pm$0.22                  &                  89.58$\pm$0.12                  &                  93.88$\pm$0.14                  & 52.30$\pm$0.13 & 72.05$\pm$0.21 & 78.32$\pm$0.20 &30.12$\pm$0.04 & 16.08$\pm$2.85 & 44.00$\pm$2.63 & 58.61$\pm$2.70 \\
vertical flip &                  50.44$\pm$0.74                  &                  70.77$\pm$0.59                  &                  78.00$\pm$0.36                  & 36.25$\pm$0.33 & 52.57$\pm$0.42 & 58.32$\pm$0.32 & 20.75$\pm$0.11 & 13.04$\pm$1.88 & 40.35$\pm$3.18 & 56.17$\pm$3.36 \\
rotation &                  44.34$\pm$0.69                  &                  68.38$\pm$0.44                  &                  76.83$\pm$0.50                  & 32.84$\pm$0.66 & 50.14$\pm$0.48 & 57.14$\pm$0.73 & 17.94$\pm$0.21 & 6.35$\pm$2.39 & 17.48$\pm$2.64 & 27.92$\pm$3.19 \\
\hline
None & 72.06 & 90.1 & 94.02 & 53.12 & 73.14 & 78.94 & 30.62 & 20.00 &  53.91 & 68.70 \\
\hline
\end{tabular}
}
\end{center}

\end{table*}

\begin{figure*}[!t]
	\centering
	\subfigure[Recognition on TU-Berlin]{
		\label{fig:recognition-tuberlin-boxplot-acc1}
		\includegraphics[width=0.32\textwidth]{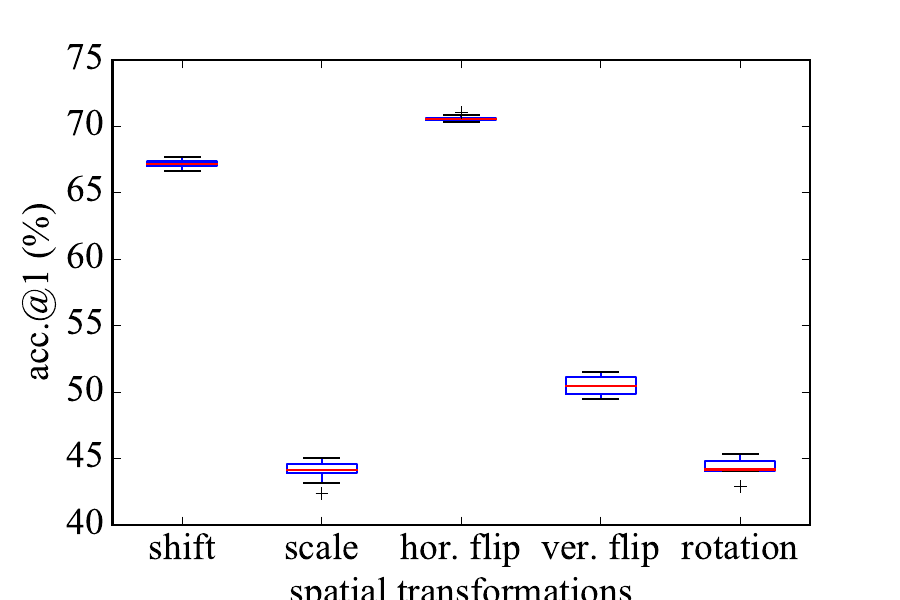}}
	\subfigure[SBIR on TU-Berlin Extended]{
		\label{fig:sbir-tuberlin-boxplot-mAP}
		\includegraphics[width=0.32\textwidth]{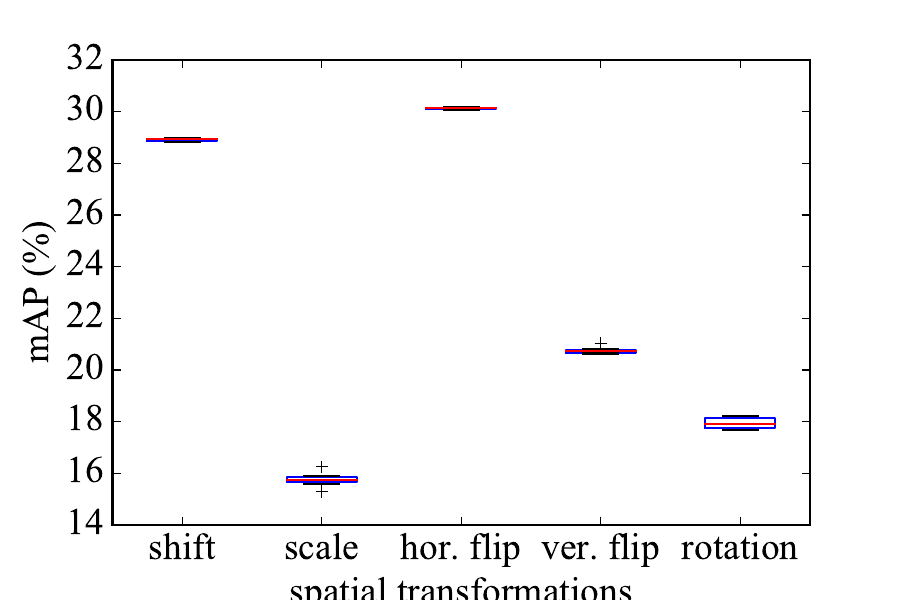}}
	\subfigure[Fine-Grained SBIR on QMUL Shoe]{
		\label{fig:sbir-shoe-boxplot-rank1}
		\includegraphics[width=0.32\textwidth]{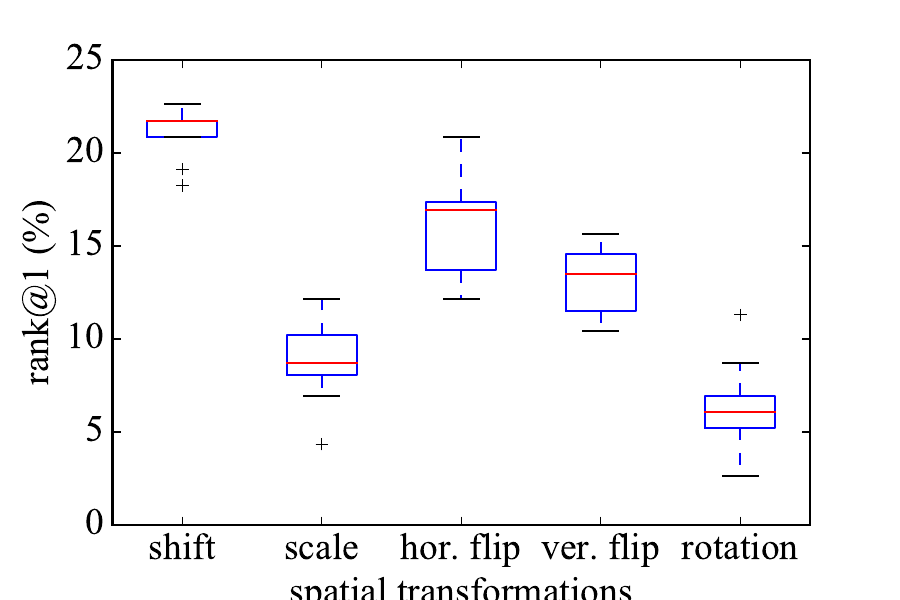}}
	\caption{{Boxplots for robustness evaluation on spatial transformations.}}
	\label{fig:boxplot}
\end{figure*}

\subsubsection{{Robustness Study on Spatial Transformation}}
\label{sec:robustness-study}
{As discussed in Section~\ref{sec:Intrinsic_Traits_and_Domain-Unique_Challenges}, even if sketches are shifted, rescaled, rotated, or flipped, they still can be recognized easily by people. It will be interesting to evaluate how sensitive the current deep networks are to the spatial transformations on different sketch tasks and datasets.}

{ To facilitate comparison, we choose three quantitatively comparable tasks as target tasks involving both single-modal and multi-modal settings, \ie, sketch recognition, coarse-grained SBIR, fine-grained SBIR, which will be respectively conducted on three commonly used datasets, \ie, TU-Berlin, TU-Berlin Extended, QMUL Shoe. To make it more intuitive, we choose a CNN backbone, \ie, ResNet-18.}

{Cross-entropy loss and triplet loss are used for recognition and SBIR tasks, respectively. To perform early-stopping and select models based on validation performance, we need to split the datasets: (i) For TU-Berlin, $40$, $20$, and $20$ sketches per category are randomly selected for training, validation, and testing, respectively. (ii) For TU-Berlin Extended, both sketches and photos are randomly divided into training, validation, and testing sets as a ratio of $2:1:1$. (iii) For QMUL Shoe, $50$ sketch-photo pairs are selected randomly from its training set for validation. To fully verify the sensitivity, we did not adopt any data augmentations in the training stage.}

{We choose five representative randomly spatial transformations for robustness testing, including position shift, scale, horizontal flip ($p=0.5$), vertical flip ($p=0.5$), and center rotation ($-45^{\circ}$ to $45^{\circ}$). Considering the randomness, testing was repeated $10$ times with each spatial transformation, and both mean ($\%$) and standard deviation for each indicator are reported in Table~\ref{table:robustness}, where top-K accuracy and rank accuracy are used as metrics for recognition and retrieval. Moreover, mean average precision (mAP) is also reported to evaluate the performance for coarse-grained SBIR as multiple true matches are provided by gallery. For a clear comparison, the testing results without any spatial transformations are provided in the bottom row of Table~\ref{table:robustness}, while we draw boxplots (Figure~\ref{fig:boxplot}) based on ``acc.@1'', ``mAP'', and ``rank@1'' for the selected tasks, respectively.}

{In Table~\ref{table:robustness}, we observed that: (i) Deep sketch models are vulnerable to the spatial transformations that can result in performance degradation.  (ii) Compared with shift and horizontal flip, the three other transformations cause more noticeable performance changes. In particular, for TU-Berlin sketches, horizontal flip causes very small performance changes. (iii) On QMUL Shoe, spatial transformation based perturbations have relatively large standard deviations. Moreover, it is interesting to see that shift transformation slightly improves fine-grained SBIR accuracy. This is also demonstrated by Figure~\ref{fig:boxplot}. This is likely due to the small sample size (only $115$ sketch-photo pairs for testing).} 

{These observations indicate that sketch based spatial transformations are able to attack deep networks. The current deep learning technique still needs to be improved to achieve  sketch-oriented robustness.}

%

\section{Discussion}
\label{sec:thinking_and_discussion}

\begin{table*}[!t]
\caption{{Comparison of deep learning and traditional approaches to sketch-related tasks.}}
\label{table:deep-vs-traditional}
\begin{center}
\resizebox{\textwidth}{!}{
\begin{tabular}{ l | l | l}
\hline
Method & Advantages & Disadvantages \\
\hline
\multirow{3}{*}{Deep Learning} &   {relatively} superior performance   &   manually-designed network structures \\
&  end-to-end mapping     &    more parameters/resource cost\\
& {larger model capacity to support big data}  &  {vulnerable to overly fit}  \\
&  more training data brings higher performance in general  &  less rigorous in mathematical expression and interpretability \\
\hline
\multirow{3}{*}{Traditional} & clear in mathematical expression and interpretability   &   {relatively} suboptimal performance  \\

                                                           &   fewer parameters/resource &   manually-designed features   \\

                                                           &  less over-fitting   &  {under-parameterized for big data}  \\
\hline
\end{tabular}
}
\end{center}

\end{table*}

\begin{table*}[!t]
\caption{{Comparison of different deep network architectures for sketch-oriented tasks.}}
\label{table:theoretical-comparison}
\begin{center}
\resizebox{\textwidth}{!}{
\begin{tabular}{ l | l | l | l | l | l}
\hline
Arc. & Input & Model Space & Motivations & Advantages & Disadvantages \\
\hline
\multirow{4}{*}{CNNs} &   picture & {spatial}  & imitate human reception field & good performance in global-level tasks & weak in stroke-level tasks \\
&   (full sketch)  & {(Euclidean)} & regard sketch as binary pixel matrix & {perceive full sketch (no information lost)}   & fail in  temporal tasks  \\
 &   ~  & ~ & ~ & can stack deeper layers   & {relatively more parameters}   \\
 &   ~  & ~ & ~ & allow a variety of perception granularities  & {vulnerable to overly fit}   \\
\hline
\multirow{3}{*}{RNNs} &   stroke vector &  {spatial\&temporal} & imitate human sketching process & recurrently capture stroke temporal patterns  & {weak in stacking deeper layers} \\
&   (key points)  & ~ & ~ & {relatively fewer parameters} & weak in long stroke sequences \\
&   ~ & ~ & ~ &  & perception granularity is fixed \\
\hline
\multirow{3}{*}{\tabincell{l}{GNNs + \\ {Transformers}}} &   stroke vector  & {spatial\&temporal} & represent sketch as graph & higher flexibility in network design &  {weak in stacking deeper layers} \\
&    (key points) & ~ & (key point $\rightarrow$ node, & {relatively fewer parameters} & abstract network structure \\
&   ~ & ~ & stroke $\rightarrow$ node, \etc) &  can handle long stroke sequences &   \\
\hline
\multirow{3}{*}{TCNs} &   stroke vector  &  {spatial\&temporal} & imitate temporal reception field & concise network architecture & {weak in stacking deeper layers}  \\
&    (key points) & ~ & (key point $\rightarrow$ word, & {relatively fewer parameters} & fail to  recurrently perceive strokes  \\
&   ~ & ~ & stroke $\rightarrow$ sentence) & allow a variety of perception granularities &   \\
\hline
\end{tabular}
}
\end{center}

\end{table*}

\subsection{Open Problems}
\label{sec:open_problems}

\subsubsection{{Deep Learning vs. Traditional Methods}}
{
In recent years, deep learning methods have achieved the \sota~in all the sketch tasks.
However, some open problems still need further study, \eg,
(i) What are the advantages and disadvantages of deep learning and traditional methods on sketch?
(ii) Why do deep learning networks work well on sketch?
(iii) How are sketch-unique characteristics modelled by various deep learning networks?
}

{
Generally traditional methods work in two stages, \ie, feature engineering from sketch space to feature space, mapping from feature space to target space, while generally deep learning methods do end-to-end mapping directly from sketch space to target space. Thus the essential opportunity for exploiting human insight is in hand-designing network structures vs. hand-designing features.
Due to the sketch domain challenges (\eg, abstract, noisy, sparse, diverse), the difficulty of feature engineering for sketch is one of the main bottlenecks of the traditional methods. This is somewhat ameliorated by deep learning methods which have the capacity to learn strong feature representations given sufficient data to model their variability and diversity. 
Furthermore, deep learning networks have various motivations and mechanisms to handle the sketch-unique characteristics:
(i) CNNs use convolution filters to imitate the human reception field, and treat sketch as a binary matrix in pixel space.
(ii) RNNs imitate the temporally-extended human sketching process to recurrently capture {spatial and temporal patterns of stroke},
(iii) GNNs represent sketches as graphs, and can encode both topological{/spatial} and temporal patterns of stroke sequences,
and (iv) TCNs imitate a temporal reception field on stroke sequence.
\cut{Based on the current techniques, RNN, GNN, and TCN need to truncate or pad their input sketch strokes to the unified length. Truncating strokes will lose information.}

We summarize the advantages and disadvantages of deep learning and traditional methods on sketch in Table~\ref{table:deep-vs-traditional}; and we compare different network architectures for sketch oriented tasks in Table~\ref{table:theoretical-comparison}.

}

\subsubsection{Data and Annotations}
\keypoint{Annotation} Uni-modal sketch datasets have begun to provide some annotation such as grouping \cite{li2019toward} and segmentation \cite{qi2019sketchsegnet+}. However more fine-grained annotation of this type is necessary, wit sketch attributes in particular being lacking. As discussed in   Section~\ref{sec:sketch_photo_generation}, existing multi-modal sketch benchmark annotation is primarily in terms of  \emph{pairings} (\eg, sketch-photo, sketch-3D). However, fine-grained/local annotations (such as stroke-contour, parts, and  attributes) would enable richer cross-modal alignment models to be learned. 


\keypoint{Meta-Data and Fairness} Unlike photos, sketches are uniquely influenced by the demographics, perception, memory, and drawing style of the artist; as well as the conditions under which they are drawn (\ie, time-limited or not). Most existing datasets do not take care to acquire balanced samples of users across background, age, gender, \etc; or record such meta-data about their participants. However, such sampling and meta-data are necessary for studying changes in drawing style with these covariates, as well as for ensuring that sketch-based applications work well for users of different backgrounds. Furthermore, existing sketch datasets are mainly created as bespoke efforts by researchers or casual participants in online games with time-pressure. These may lead to sketches that are either excessively well dawn, or too poorly drawn. Data collected under a variety of drawing conditions, and meta-data about those conditions, would help sketch research in future. 


\subsubsection{Architectures and Sketch-Specific Design}
\keypoint{Architectures} As discussed in Section~\ref{sec:comparison} and Table~\ref{table:comparison_on_recognition}, there are a variety of network architectures that can be applied to sketch, and these lead to a range of performances. These results suggest that performance will continue to advance as better architectures within each family are developed. The best architecture for sketch perception is still an open question. 

\keypoint{Sketch-Specific Design} Another open question is to what extent sketch-specific designs are important vs.~generic computer vision architectures and learning algorithms. Clearly sketch has unique challenges (sequential time-series nature, sparsity, abstraction, artist style, \etc) that can be better taken into account with sketch-specific designs. However the broader vision and learning community can bring greater effort to bear on developing more advanced general purpose models. Therefore it remains to be seen in which sketch applications sketch-specific designs can take a decisive lead over generic architectures and algorithms. For example, sketch-specific designs may be more important in fine-grained tasks such as segmentation, grouping, and FG-SBIR compared to the simplest coarse-grained object categorization task.

\keypoint{{Association with Other Sketches}} Some CNN based models designed for other kinds of sketches (\eg, well-drawn line drawings~\cite{simo2016learning,favreau2016fidelity,sasaki2017joint,simo2018mastering}, cartoons~\cite{simo2016learning,simo2018mastering}) can be applied to free-hand sketch. In particular, it has been verified by relevant literature that CNN models oriented at high-quality line drawings can be successful for free-hand sketch tasks, including vectorization~\cite{kim2018semantic,bessmeltsev2019vectorization,guo2019deep,HaoranSIGGRAPH2021}, shading~\cite{venkataramaiyer2021shad3s}, inking~\cite{SimoSerraSIGGRAPH2018}, \etc. It is interesting to evaluate which methods designed for other sketches can or cannot be applied to free-hand sketches. This will depend on to what extent architectures designed for well-drawn sketches (Figure~\ref{fig:out-of-scope}) become over-specialized to that type of data, or can generalize to the more abstract, diverse, and sparse free-hand sketches (Figure~\ref{fig:sketches}). 


\subsection{Potential Research Directions}
\label{sec:future_directions}
In this section, we outline some potential research directions that we believe are promising in future, {from both the perspectives of potential application and underpinning research value}.

\subsubsection{{Potential Application-Oriented Research}}


\keypoint{{Scene Sketches}} 
{
Scene-level sketch oriented deep learning is still under-studied.
Recently, several seminal works (\eg, SketchyCOCO~\cite{Gao2020SketchyCOCO}, SceneSketcher~\cite{liu2020scenesketcher}) have opened up   promising directions for scene-level sketch research. They  not only contribute large-scale scene-level sketch datasets but also propose novel research topics, \eg, scene sketch based image generation~\cite{Gao2020SketchyCOCO}, fine-grained scene sketch based image retrieval~\cite{liu2020scenesketcher}.
These novel topics are useful for the practical applications, \eg, sketch based scene design. It remains to be seen up to  complexity of scenes are technically feasible - and practically appealing for users - to retrieve using sketches. 
}

\keypoint{{3D Sketches}} 
{
Collecting 3D sketches~\cite{xu2018model} is now easier thanks to new data collection equipment. This could support many interesting 3D sketch related research topics,
 \eg, combining virtual reality (VR)~\cite{jackson2016lift} and augmented reality (AR)~\cite{giunchi2019mixing,kwan2019mobi3dsketch,gasques2019you}. 
 3D sketch research will help to bring sketch-based human-computer interaction from the 2D plane of the touch-screens to 3D spaces, enabling more immersive experience.
}

\keypoint{Diverse Sketch Subjects} Existing sketch datasets and applications mainly focus on sketches depicting objects. However, in practical applications users may be interested in machines understanding more diverse sketched concepts, \eg, sheets, curves, histograms~\cite{roberts2015sketching}, maps~\cite{boniardi2016autonomous}, engineering sketches~\cite{Willis_2021_CVPR}, and user interface (UI) prototype drawings~\cite{jain2019sketch2code}.
Sketch also can be studied together with hand-written characters.

\keypoint{Sketch Color and Pressure} Existing free-hand sketches are collected by common touch-screen devices, \eg, phone or  tablet. Here the position of strokes is the main feature, with color and texture not being widely collected. Thus, existing sketch analysis has mostly focused on black or grayscale sketches. However, sketches can already be colored, and recent devices can increasingly sense pressure along strokes. Upgrading models to exploit color, texture, and pressure properties of sketches remains as outstanding work. 


\keypoint{{Sketch Beautification}} {How to beautify sketch~\cite{paulson2008paleosketch} in various sketch-related HCI applications is interesting and challenging. Sketch beautification will not  only refine user experience but also improve interaction efficiency in the sketch-based design applications, \eg, modifying a non-professional sketch to a professional sketch~\cite{li2020sketchman}.}

\keypoint{{Sketch-Based Design}} {People often use very simple or even scrawled sketches at the beginning of a design to brainstorm and generate inspiration. Thus, there are some very useful sketch-based designing applications that can help people to design 2D or 3D products, \eg, sketch-to-comic~\cite{sykora2009lazybrush,sykora2011textoons,liu2018auto,kim2019tag2pix}, sketch-to-shadow~\cite{zheng2020learning,venkataramaiyer2021shad3s}, and sketch-to-normal~\cite{hudon2018deep,su2018interactive}. We believe that in future these techniques can be continually improved further by future deep learning techniques. These sketch-based designing methods can improve the efficiency in HCI. 

\keypoint{{Efficient Models}} For sketch-based perception and user-interfaces in mobile devices such as phones, tablets, or AR/VR gear, processing should be real-time and light-weight enough to run on embedded battery powered devices. How to compress sketch models and improve their efficiency is an important question for future work.}

\subsubsection{{Potential Theoretical Research}}

{
Sketch-oriented deep learning models have achieved good performance within well curated datasets. However, how well they can generalise to uncontrolled real-world conditions remains to be seen. 
}

\keypoint{Diversity, Style, and Robustness} The impact of dataset shift~\cite{Yesilbek_2021_CVPR} has only begun to be studied in sketches. More uniquely, sketch is particularly subjective in terms of influence by the user's drawing style, culture and potentially demographics. Robustness to these stylistic aspects is an under-studied area of sketch research. 
{Similarly, taking the native format of sketch, existing adversarial attack studies could be extended to sketch domain by studying adversarial \emph{strokes} or \emph{waypoints}, rather than pixel perturbations.}

\keypoint{Sketch as a Robustness Test} Sketch images differ dramatically from photos, yet are easily recognized by humans. 
{Sketch images such as ImageNet-Sketch~\cite{wang2019learning}, SketchTransfer \cite{lamb2020sketchtransfer}, and PACS \cite{Li_2017_ICCV} can thus be used to benchmark the robustness and generalization ability of more general image-oriented deep learning models. Going beyond recognition, similar robustness evaluations could be performed on other instance-level retrieval problems such as person Re-ID.}

\keypoint{Data-Efficient Sketch Models} Major process in deep learning for sketch research has been driven by increasingly large sketch datasets (Table~\ref{table:dataset_table}). However, ultimately these datasets are harder to scale than corresponding photo datasets due to the need for manual sketching. Therefore data-efficient approaches to all the main sketch-analysis tasks of interest from recognition to SBIR are of major importance going forward. Whether this is best achieved by few-shot learning, self-supervised learning, or cross-modality knowledge transfer from photo domain remains to be seen. 


%


\section{Conclusion}
\label{sec:conclusion}
This survey reviewed the landscape of contemporary deep-learning based sketch research. We introduced the unique aspects of sketch in terms of sketch-specific challenges and diverse potential representations, and analyzed both existing datasets and existing methods in terms of a rich ecosystem of uni-modal and multi-modal sketch analysis tasks. We discussed open problems and under-studied research directions throughout. 
We hope this survey will help new researchers and practitioners get up to speed, provide a convenient reference for sketch experts, and encourage future progress in this exciting field.

\ifCLASSOPTIONcaptionsoff
  \newpage
\fi

\bibliographystyle{IEEEtran}
\bibliography{clean_egbib}

\end{document}